\documentclass{article}

\usepackage[preprint]{corl_2026} %

\usepackage{booktabs}
\usepackage{multirow}
\usepackage{graphicx}
\usepackage{xcolor}
\usepackage{amsmath}
\usepackage{amssymb}
\usepackage[acronym]{glossaries}
\usepackage{algpseudocode}
\usepackage{tabularx}
\usepackage[table]{xcolor}
\usepackage{subcaption}
\usepackage{graphicx}
\usepackage{array}
\usepackage{tabularx}
\usepackage{booktabs}
\usepackage{graphicx}
\usepackage{rotating}
\usepackage{float}
\usepackage{lipsum}
\usepackage{cleveref}

\usepackage{graphicx}
\usepackage{array}
\usepackage{tabularx}
\usepackage[export]{adjustbox}

\newcolumntype{Y}{>{\centering\arraybackslash}X}

\usepackage[T1]{fontenc}
\usepackage[scaled=0.92]{inconsolata} %
\usepackage{subcaption}
\usepackage{listings}
\usepackage{xcolor}

\lstdefinestyle{promptstyle}{
  basicstyle=\ttfamily\tiny,
  breaklines=true,
  columns=fullflexible,
  keepspaces=true,
  frame=single,
  framerule=0.3pt,
  rulecolor=\color{black!30},
  aboveskip=0.3em,
  belowskip=0.3em,
  xleftmargin=0.3em,
  xrightmargin=0.3em,
  framesep=3pt
}

\usepackage{booktabs}
\usepackage[table]{xcolor}
\usepackage{amssymb}
\usepackage{array}
\usepackage{wrapfig}
\usepackage[table]{xcolor}
\usepackage{pifont}
\usepackage{array}
\usepackage{array}
\newcolumntype{C}[1]{>{\centering\arraybackslash}p{#1}}
\newcommand{\cmark}{\textcolor{green!55!black}{\ding{51}}}
\newcommand{\xmark}{\textcolor{red!70!black}{\ding{55}}}

\definecolor{oursgray}{gray}{0.92}

\usepackage{makecell}
\usepackage[table]{xcolor}

\usepackage{amssymb}
\usepackage[table]{xcolor}

\usepackage{bm}

\glsdisablehyper
\newacronym{lfd}{LfD}{Learning from Human Demonstration}
\newacronym{sota}{SoTA}{state-of-the-art}
\newacronym{method}{SPARC}{Spatial Annotations from Robot Demonstrations with Reliability Calibration}
\newacronym{bench}{IA-Bench}{Interaction-Aware Bench    }
\newacronym{vla}{VLA}{Vision-Language-Action Model}
\newacronym{vlm}{VLM}{Vision-Language Model}
\newacronym{bsp}{B-Spline}{B-Spline}
\newacronym{cbsp}{clamped B-Spline}{clamped B-Spline}
\newacronym{dof}{DoF}{Degrees of Freedom}
\newacronym{iab}{IA-Bench}{IA-Bench}

\title{SPARC: Reliable Spatial Annotations from Robot Demonstrations at Scale}

\author{
  Nils Blank$^{1}$ \quad Paul Mattes$^{1}$ \quad Maximilian Xiling Li$^{1}$ \quad Jakub Suliga$^{1}$ \quad Thomas Roth$^{1}$ \\
  \textbf{Moritz Reuss$^{2}$ \quad Pankhuri Vanjani$^{1}$ \quad Rudolf Lioutikov$^{1,3}$} \\[0.3em]
  {\normalfont $^{1}$Karlsruhe Institute of Technology \quad $^{2}$NVIDIA \quad $^{3}$Robotics Institute Germany}
}

\begin{document}
\maketitle

\begin{abstract}
This work introduces \gls{method}, a risk-aware framework that automatically labels robot demonstrations with structured spatial annotations and assigns each annotation a reliability score.
Such annotations, including bounding boxes, object trajectories, and manipulation phase labels, support grounded robot policies, embodied foundation models, and motion planning.
Existing automated pipelines generate such annotations at scale but provide no reliable quality signal: detector confidence is poorly calibrated for annotation correctness, forcing a choice between accepting noisy labels or discarding useful samples.
In contrast, \gls{method} predicts annotation correctness from physical interaction evidence in the demonstration, such as phase-aligned object motion, gripper proximity, and overlap with the robot, thus reducing noisy labels and retaining more useful samples.
We find that these cues play complementary roles. No single cue both selects the manipulated object and identifies which annotations to discard.
We further introduce \gls{bench}, a benchmark that measures model accuracy in grounding the locations of interacted 
objects in robot demonstrations. On $1.7$k human-annotated demonstrations spanning 
diverse embodiments and scenarios, SPARC significantly 
outperforms detection-only baselines in object localization accuracy while also retaining three times more samples at high-precision operating points. 
Our experiments demonstrate that models finetuned on our annotations achieve state-of-the-art results on object-grounding and pointing benchmarks among similarly sized models, while remaining competitive on broader spatial-reasoning suites without any manually verified or annotated training data. Furthermore, 
policies trained on SPARC-generated annotations significantly outperform baselines in cluttered, visually ambiguous real-world scenes. Code, data, and models are publicly available at https://sparc-labeling.github.io. 
\end{abstract}

\begin{figure*}[h!]
    \centering
    \includegraphics[width=1\linewidth]{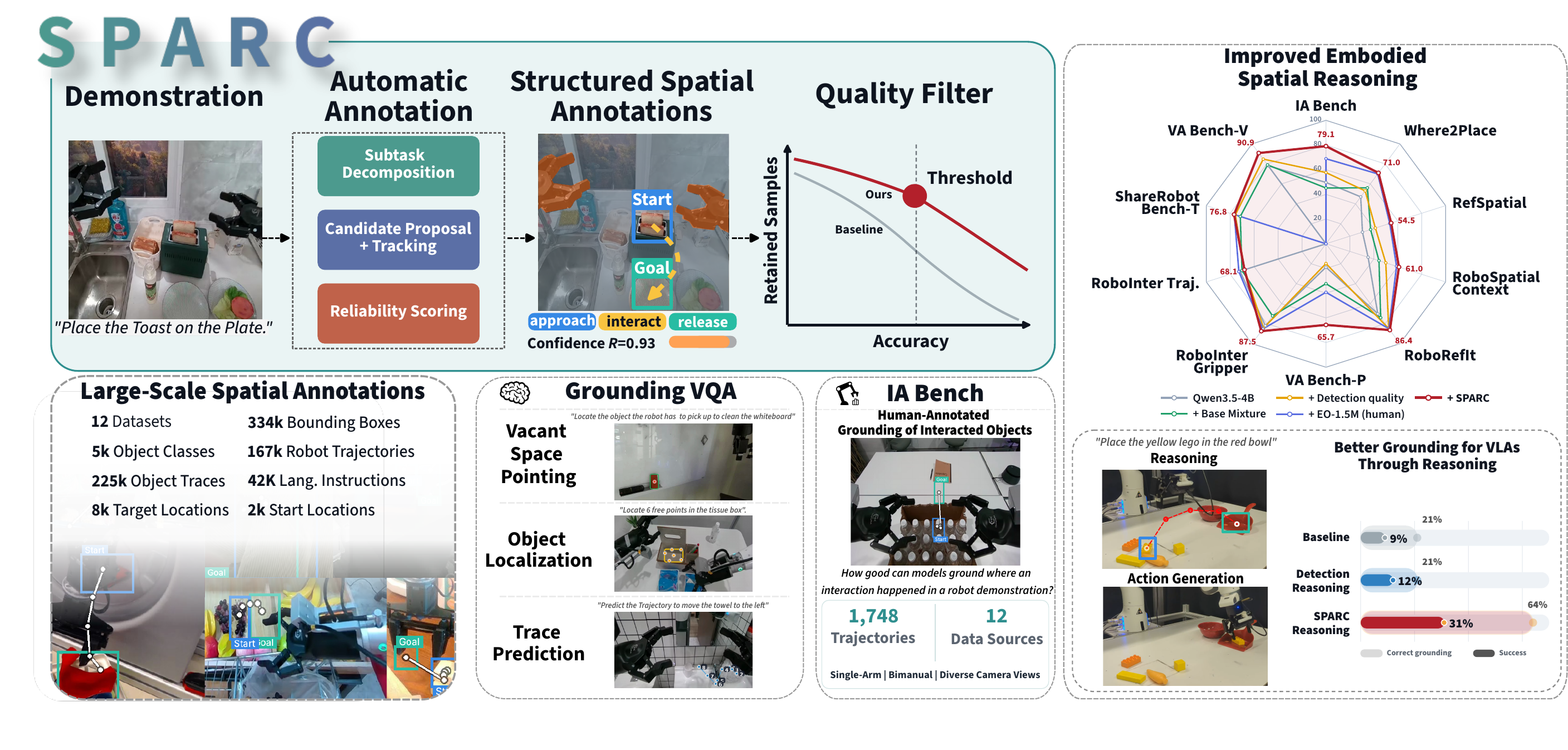}
\caption{\textbf{SPARC} auto-labels robot demonstrations with object-centric spatial annotations and a per-annotation reliability score derived from interaction evidence: phase-aware motion, gripper proximity, and a robot-overlap filter. A single threshold on this score controls the quality-coverage tradeoff without human review, producing large-scale annotations (\emph{bottom}) that improve downstream embodied reasoning and policy learning (\emph{right}).}
    \label{fig:teaser}
\end{figure*}

\section{Introduction}
\label{sec:intro}

Spatial annotations of robot demonstrations identify which object is being manipulated, where it is located, and how it moves through time. These annotations are a key ingredient in modern robot learning pipelines, including object-centric reasoning in \glspl{vla}~\citep{zawalski2024ecot, zhao2025cotvla, bai2026latent}, pixel-level scene understanding for visually grounded policies~\citep{liang2024pixelvla}, and embodied \gls{vlm} mid-training~\citep{zhang2026vlmvla, embodiedmidtrain_2026, robobrain2025, molmoact2, dang2026rynnbrain}. 

Producing such annotations reliably at scale remains difficult.
Real-world demonstrations contain clutter, occlusion, and large viewpoint variation. 
Existing automated pipelines typically chain a detector with a tracker~\citep{zawalski2024ecot, bai2026latent, liang2024pixelvla, Shou2026HALOAU}, making the detector the sole source of object identity. Detector confidence measures recognition confidence, not whether the detected object is the one actually being manipulated. 
In cluttered scenes, such pipelines can therefore lock onto a plausible but incorrect object with high confidence and propagate that error through time.
This mismatch creates a direct scale-quality tradeoff. Hard thresholds on detector confidence either retain many incorrect annotations or discard many valid ones. 
Human review at annotation time is reliable but expensive \cite{yang2025sam2auto, lirobointer, qu2025eo, li2025hamster, robobrain2025}. Simulation provides ground-truth \cite{yuanrobopoint, zhou2026roborefer} but struggles with real world transfer.

To address these challenges, we introduce \textbf{Sp}atial \textbf{A}nnotations from \textbf{R}obot Demonstrations with Reliability \textbf{C}alibration 
(\textbf{\gls{method}}), an automatic pipeline that produces spatio-temporal annotations from robot demonstrations. 
For each object-centric subtask, \gls{method} identifies the interacted object and generates its start and target locations, its movement over time as a 3D object trajectory, and a continuous reliability score. 
\gls{method} builds on off-the-shelf perception models. Its core is a \textbf{reliability score} that predicts annotation correctness from physical interaction evidence rather than appearance.
Phase-aligned motion and 3D gripper proximity identify the manipulated object, overlap with the robot's own geometry provides negative evidence, and detector confidence anchors the score across demonstrations.
The resulting per-annotation score lets users explicitly trade off annotation quality and dataset scale without human review.

To evaluate \gls{method}, we introduce Interaction-Aware Bench (IA-Bench), a novel benchmark of 1.7k hand-annotated start and target locations of manipulated objects spanning 12 diverse embodiments and settings, including bimanual setups. 
Existing spatial annotation benchmarks cover only single-frame observations, making them insufficient to assess interacted object identity and its spatio-temporal properties. Our full pipeline achieves 80.2\% interacted object localization accuracy at high confidence, compared to 58.1\% for a detection-only baseline, while remaining viable under aggressive quality filtering where detection-only pipelines collapse to near-zero usable data.

We then show that \gls{method} annotations are useful downstream.
Annotating 280k robot trajectories and thresholding for 95\% precision yields 167k high-quality trajectories, versus only 30k from detector confidence at the same target, from which we construct 511k VQA samples for vacant-space pointing, trace prediction, and object grounding.
Although none of our annotations are human-verified, models fine-tuned on this data outperform similarly sized models trained on human-annotated data on object-grounding and pointing benchmarks~\cite{yuanrobopoint, yuan2025fsd, lu2023vl, zhou2026roborefer, robobrain2025}, and reasoning policies trained on them outperform detection-annotated ones in cluttered real-world manipulation.

\textbf{Contributions.}
(i) We frame automatic spatio-temporal annotation of robot demonstrations as a \emph{selective-prediction} problem and introduce a per-annotation reliability score grounded in physical interaction evidence. We show that interaction cues play distinct roles as selectors, confidence anchors, and negative evidence, that no single cue achieves both selection and abstention, and that these roles transfer to unseen embodiments and egocentric human videos.
(ii) We introduce \gls{bench}, an interaction-aware benchmark with video and proprioception spanning diverse embodiments and bimanual setups.
(iii) We generate a large-scale, high-quality dataset of 167k annotated trajectories and show that fully automatic \gls{method} supervision matches or exceeds human-annotated data on embodied grounding and yields reasoning policies more robust in cluttered real-world manipulation.

\section{Related Work}
\label{sec:related}

\textbf{Automatic spatial annotation for embodied data.}
Generating spatio-temporal supervision for robotic manipulation is crucial for embodied learning. Existing pipelines typically localize grippers and task-relevant objects using frame-by-frame detectors~\cite{zawalski2024ecot, chen2025training, liang2024pixelvla, huang2026fast, huang2026thinkact, lee2025molmoact}, often augmented by temporal models for object tracking~\cite{yuan2025fsd, bai2026latent, Shou2026HALOAU, gan2025foundationmotion, liu2026moright} or proprioceptive data for language generation~\cite{chen2026robovlm, bai2026latent, chen2025training}. To counteract tracking failures, existing frameworks rely on expensive human verification~\cite{qu2025eo, lirobointer, robobrain2025, li2025hamster, team2025gemini, dang2026rynnbrain} or simulation environments~\cite{yuanrobopoint, chen2025internvla, denggraspvla, li2025multiobjective, mattes2026sir}, though the latter suffers from reality-gap discrepancies. Egocentric human-object interaction datasets~\cite{liu2022hoi4d, darkhalil2022epic, perrett2025hd, guo2023handal, hoque2025egodex, lazarow2025cubify} offer alternative supervision but differ in viewpoint and embodiment. Like prior pipelines, \gls{method} builds on off-the-shelf detection, tracking, and depth models. It differs in using temporal and proprioceptive structure to verify which candidate was actually manipulated, and in attaching a reliability score to every annotation.

\textbf{Reliability scoring and data filtering.}
Data filtering via continuous reliability metrics is a well-established paradigm across semi-supervised learning~\cite{sohn2020fixmatch, xu2021end, huang2019mask, northcutt2021confident}, automated segmentation~\cite{kirillov2023segment, ravi2025sam, carion2025sam}, vision-language pre-training~\cite{datacomp, fang2024data, chen2024alpagasus, zhou2023lima}, and point tracking~\cite{karaev2025cotracker3}. However, these confidence scores transfer poorly to robotic manipulation due to severe occlusions and gripper-object overlaps. Consequently, current robotic pipelines rely primarily on binary heuristics, thresholding based on per-frame detector confidence, or treating it as an initial source of truth. This reflects category recognition rather than physical manipulation~\cite{zawalski2024ecot, liang2024pixelvla, chen2025training}. Alternatively, recent works filter by motion magnitude, which discards static interactions~\cite{yuan2025fsd, bai2026latent} and still relies on an initial object detector. \gls{method} addresses this gap by aggregating multi-modal spatial and temporal interaction cues into a continuous per-annotation reliability score, enabling fine-grained control over the quality-coverage tradeoff without human intervention. Unlike frameworks that generate text-based language annotations without spatial tracking~\cite{blankscaling} or score data only at a coarse, whole-demonstration level~\cite{zhang2025scizor, chen2025curating, jangdreamgen}, \gls{method} produces and scores precise spatio-temporal annotations.

\section{Method}
\label{sec:method}

\begin{figure}[h]
    \centering
    \includegraphics[width=\linewidth]{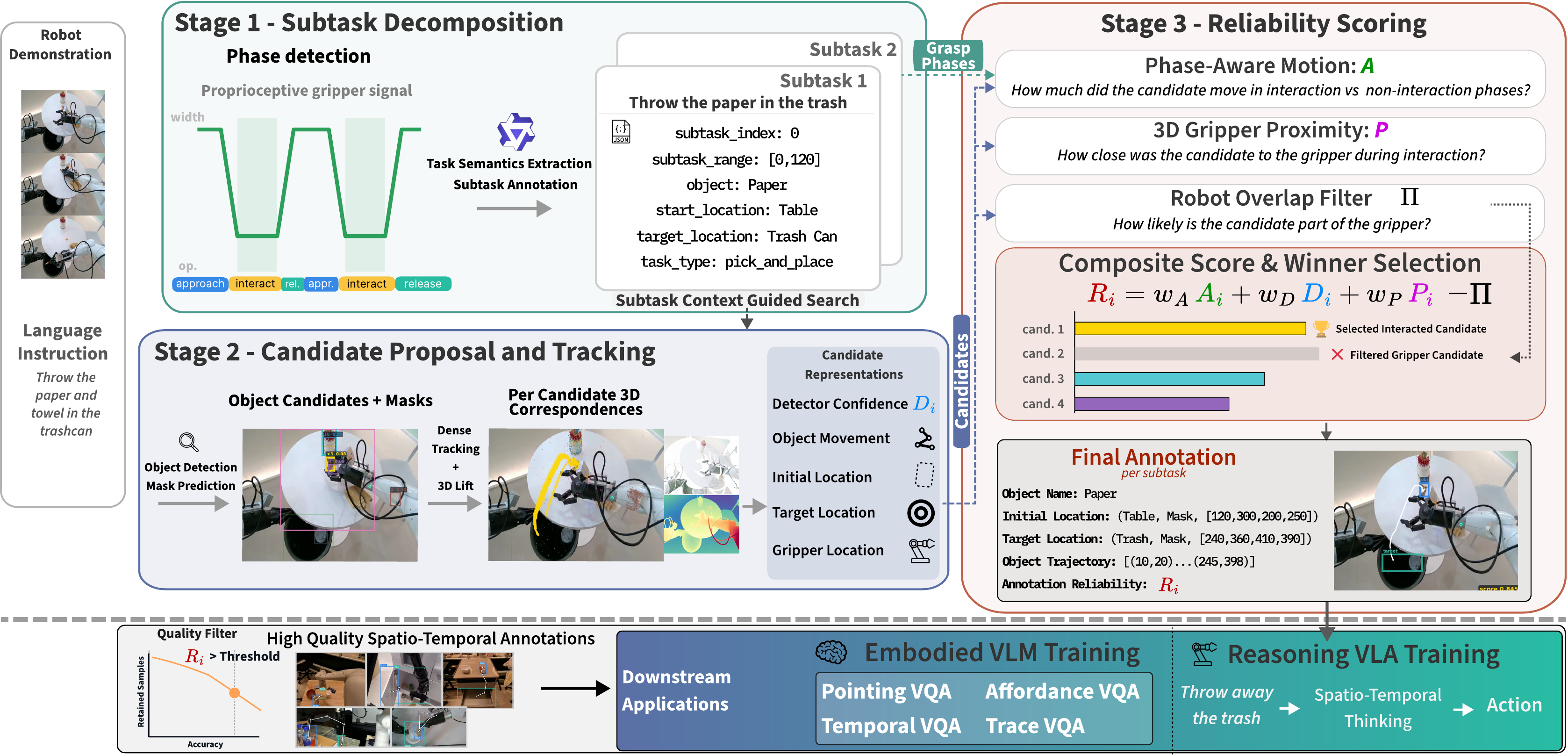}
\caption{\textbf{Overview of the SPARC annotation pipeline.}
Stage~1 segments a demonstration into object-centric subtasks via gripper-phase detection and
language parsing. Stage~2 proposes, tracks, and 3D-lifts object candidates. Stage~3 scores each
candidate with phase-aware motion $A$, 3D gripper proximity $P$, and a robot-overlap filter,
combined with detector confidence $D$ into a composite reliability score $R$. The top-scoring
candidate yields the annotation, and thresholding $R$ controls the quality-coverage tradeoff
without human review.}
    \label{fig:pipeline}
\end{figure}

Given a robot demonstration video, a language instruction, and gripper proprioceptive signal,
\gls{method} produces object-centric spatial annotations for each interaction segment.
Each annotation contains a subtask instruction, the manipulated object name, initial and
target object locations, an object trajectory, gripper phase boundaries, and a scalar
reliability score.
\gls{method} proceeds in three stages (Figure~\ref{fig:pipeline}): subtask decomposition, candidate proposal and tracking, and reliability scoring.

\subsection{Stage 1 - Subtask Decomposition and Phase Detection}

\textbf{Interaction segmentation.}
Given a long-horizon demonstration \(\tau=\{(x_t,q_t)\}_{t=1}^{T}\), with visual frames \(x_t\), proprioceptive states \(q_t\), and interacted objects \(\mathcal{O}_\tau\), \gls{method} decomposes \(\tau\) into short object-centric segments $\mathcal{S}(\tau)=\{s_k\}_{k=1}^{K}.$
Each segment \(s_k\) captures a single interaction with one object \(o_k \in \mathcal{O}_\tau\). 
We then use the language instruction \(\ell\) to assign an object-centric subtask description \(u_k\) to each segment. 
For tool-use tasks, we treat the manipulated tool as the primary interaction object and also track the target of the tool use.

\textbf{Gripper phase extraction.}
Following grasp-phase detection~\citep{chen2026robovlm}, we extract maximal closed-gripper intervals $[s_k,e_k]$ from the normalized gripper signal $a_t$ using $a_t<\tau_c$. 
Short closures are removed with an adaptive duration threshold, retaining only intervals with $d_k \ge \alpha\,\mathrm{median}(\{d_j\}_{j=1}^K)$. 
We apply this independently per arm and use adjacent open-gripper intervals as approach and release phases, establishing the grasp cycle for subsequent candidate scoring.

\textbf{Language-based subtask parsing.}
To recover which object each interaction involves and how to describe it, we use Qwen3.6-30B~\citep{qwen35blog} to align the detected grasp cycles with the language instruction.
The model receives the original instruction and the ordered list of detected grasp cycles, with their approach, interaction, and release intervals.
It returns one JSON record per subtask with the object, its start and target locations, a free-form action label, optional tool use, and a description of each phase (prompts in Appendix~\ref{fig:appendix_prompts_compact}).
The output of this stage is a set of subtask records
$
    \mathcal{S}
    =
    \{(u_j, \ell o_j, q^{\mathrm{init}}_j, q^{\mathrm{target}}_j,
    \mathcal{P}_j)\}_{j=1}^{M},
$
where $u_j$ is the subtask instruction, $\ell o_j$ is the candidate object name, $q^{\mathrm{init}}_j$ and $q^{\mathrm{target}}_j$ describe the initial and target locations, and $\mathcal{P}_j$ contains the approach, interaction, and release phase intervals.

\subsection{Stage 2 - Candidate Proposal and Tracking}
\gls{method} compiles a set of candidate objects and lifts their motion to 3D.
We run LLMDet~\cite{fu2025llmdet} at a single keyframe $t^\star$, chosen halfway through the first grasp phase to reduce occlusion between the interacted object and gripper. 
We use a very low detection threshold, effectively treating LLMDet as a region proposal model while retaining its grounding confidence scores, yielding between 5 and 25 object proposals. 
For each proposal, \gls{method} obtains an instance mask with SAM2~\cite{ravi2025sam}. 
We denote the resulting candidate set $\mathcal{C}$, where each candidate
$c_i = (b_i, m_i, \ell o_i, D_i)$ consists of a bounding box $b_i$,
segmentation mask $m_i$, object label $\ell o_i$, and detector confidence $D_i$.

To capture object motion, \gls{method} applies AllTracker~\cite{harley2025alltracker},
which produces dense pixel tracks over the video. For each candidate $c_i$, we select
all tracks originating from pixels inside the object mask $m_i$ and lift them to 3D
using MoGe-2~\cite{wang2026moge}, which predicts a dense geometry map
$\Phi_t : \Omega \rightarrow \mathbb{R}^3$ per frame. The final representation for
candidate $c_i$ is $o_i = (c_i, \Gamma_i)$ with
\begin{equation}
    \Gamma_i = \left\{ \left( \bm{x}_p^t, \bm{u}_p^t, v_p^t \right) \mid p \in m_i,\;
    t \in \mathcal{T}_p \right\},
\end{equation}
where $m_i$ is the mask of object $i$, $\bm{x}_p^t$ is the 3D position of pixel $p$ at frame $t$, $\bm{u}_p^t$ its 2D image
location, and $v_p^t$ its visibility.

\begin{figure}[t]
    \centering
    \includegraphics[width=0.9\linewidth]{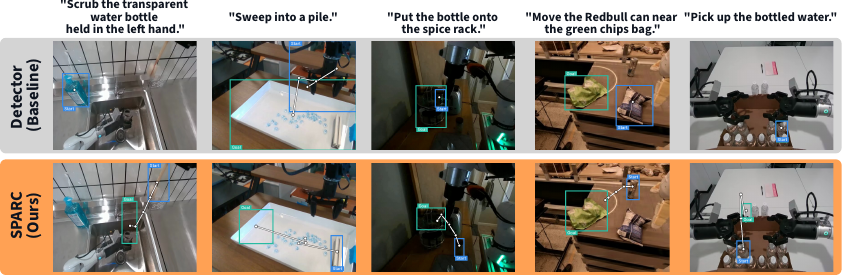}
    \caption{Qualitative comparison on diverse robot demonstrations. Columns show different language-conditioned tasks, and rows compare the detector baseline and \gls{method}. The detector often picks plausible distractors or robot parts, while \gls{method} selects the interacted object.}
    \label{fig:qualitative_mismatch_comparison}
\end{figure}
\subsection{Stage 3 - Reliability Scoring}
 
Not all tracked candidates correspond to the manipulated object, and detector confidence only captures appearance-level grounding, which aligns poorly in cluttered scenes (Figure~\ref{fig:qualitative_mismatch_comparison}).
We therefore combine cues with different roles into a reliability score $R_i$ (Sec.~\ref{sec:score_analysis}).
Phase-aware motion and 3D gripper proximity provide positive evidence of manipulation, a robot-overlap penalty provides negative evidence against the robot's own geometry, and detector confidence anchors the score across demonstrations.

\textbf{Phase-aware motion.}
Manipulated objects should move most strongly during the interaction phase. 
For candidate $i$, we compute the mean 2D displacement of its tracked object points $m_{i,t}$ at each frame.
Using the grasp-phase segmentation, we compare each object's motion during interaction intervals against its motion outside interaction intervals: 
$A_i = (m_i^{\mathrm{int}})^{\alpha} / (m_i^{\mathrm{non}} + 1)^{\beta}$ with $\alpha{=}0.6$ and $\beta{=}0.2$. 
This favors objects moving during manipulation while down-weighting background
motion and tracking jitter.

\textbf{3D gripper proximity.}
Interaction is spatially constrained by the robot embodiment: the manipulated object should be close to the gripper. 
For each frame, we construct an adaptive 3D sphere $B(g_t, r_t)$ around the gripper center $g_t$ with radius $r_t$, where $r_t$ is estimated from the local 3D gripper geometry and the extent of candidate boxes in the scene. The gripper center is estimated by the gripper point closest to the image center.
\gls{method} computes the fraction of candidate points inside the gripper sphere for each visible point $Q_{i,t}$ of candidate $i$ and aggregates over valid frames:

\begin{equation}
    p_{i,t}
    =
    \frac{1}{|Q_{i,t}|}
    \sum_{q \in Q_{i,t}}
    \mathbf{1}
    \left[
        q \in B(g_t, r_t)
    \right],
\end{equation}
and aggregate over valid frames as $P_i = \frac{1}{|\mathcal{T}_i|}\sum_{t \in \mathcal{T}_i} p_{i,t}$.

\textbf{Robot-overlap soft penalty.}
A common failure mode is selecting the robot gripper, which often moves most
and appears prominently in detection proposals.
For each candidate we compute the fraction of tracked points inside a
RobotSeg mask~\cite{mei2025robotseg} at the detection keyframe, giving an overlap
score $\in [0,1]$.
High-overlap candidates receive a soft quadratic penalty $\Pi_i$, attenuated
by their 3D depth separation from the gripper tip, so contacted objects are
retained despite visual overlap.
Further details are in Appendix~\ref{app:robot_overlap}.

\textbf{Final reliability score.}
We min-max normalize $A_i$ and $P_i$ across candidates in the same video, giving $\widehat{A}_i$ and $\widehat{P}_i$, and combine all four signals:
  \begin{equation}
      R_i
      =
      w_A\,\widehat{A}_i + w_D(\widehat{P}_i)\,D_i + w_P\,\widehat{P}_i - \Pi_i,
      \label{eq:reliability}
  \end{equation}
  where $w_A = 0.5$ and $w_P = 0.3$, and $w_D(\widehat{P}_i) = 0.75 - 0.15\,\widehat{P}_i$ is an adaptive weight that reduces reliance on appearance when 3D gripper-proximity evidence is strong.
  The selected object is $i^\star = \arg\max_i R_i$.
  To locate the target object, \gls{method} counts how many of $\Gamma_{i^\star}$'s tracks land inside each candidate target box and ranks by movement magnitude and track count.

\gls{method} annotates a trajectory in $\sim$5\,s per sample on a single GH200 GPU and parallelizes across workers for a $24\times$ wall-clock speedup over human labeling, with every artifact reused as downstream supervision so the cost is amortized (Appendix~\ref{app:runtime}).

\section{Experiments}
\label{sec:experiments}

\begin{table}[t]
\centering
\small
\setlength{\tabcolsep}{6.0pt}
\renewcommand{\arraystretch}{1.12}
\begin{tabular}{l l cccc}
\toprule
\textbf{Category} & \textbf{Scoring rule} &
\textbf{Acc.}$\uparrow$ &
\textbf{Cov@90}$\uparrow$ &
\textbf{AURC}$\downarrow$ &
\textbf{E-AURC}$\downarrow$ \\
\midrule
\multirow{3}{*}{\textit{Baselines}}
& Det. confidence ({\textcolor{black!45}{OWLv2}})   & 0.622 & 0.002 & 0.297 & 0.002 \\
& Motion magnitude                                  & 0.463 & 0.000 & 0.657 & 0.476 \\
& FSD -- Traj.\ length~\cite{yuan2025fsd}           & 0.386 & 0.331 & 0.279 & 0.032 \\
\midrule
\multirow{4}{*}{\textit{Ablation}}
& Det. confidence ({\textcolor{black!45}{LLMDet}})  & 0.581 & 0.266 & 0.219 & 0.115 \\
& \quad + Mean movement                             & 0.697 & 0.514 & 0.117 & 0.066 \\
& \quad + Robot filter                              & 0.727 & 0.549 & 0.101 & 0.059 \\
& \quad + Phase aware movement                               & 0.778 & 0.735 & 0.064 & 0.037 \\
\midrule
\multirow{2}{*}{\textit{External}}
& Gemini Robotics ER 1.6~\cite{team2025gemini}      & 0.717 & -- & -- & -- \\
& Qwen3.6-30B-A3B~\cite{qwen35blog}                 & 0.060 & -- & -- & -- \\
\midrule
\rowcolor{black!5}
& \textbf{Ours (Final)}                             & \textbf{0.802} & \textbf{0.776} & \textbf{0.056} & \textbf{0.035} \\
\bottomrule
\end{tabular}
\vspace{0.5em}
\caption{\textbf{Interaction-aware filtering ablation.}
\textit{Baselines}: prior-style confidence and motion filters. \textit{Ablation}: starting from the LLMDet detection-confidence score, we sequentially add each component. \textit{External}: general-purpose models reported for reference.}
\label{tab:main_results}
\vspace{-1.5em}
\end{table}
\textbf{Interaction-aware benchmark.}
To evaluate annotation quality, we introduce \acrfull{bench}, a benchmark of robot manipulation demonstrations with human-annotated start and target boxes for the manipulated object. \gls{bench} contains 1,748 ground-truth annotations from four data sources: AgiBotWorld~\citep{agibot2025} (451 samples), DROID~\citep{khazatsky2024droid} (405 samples), BridgeData~\citep{walke2023bridgedata} (454 samples), and Open X-Embodiment (OXE)~\citep{o2024openx} (438 samples). 
It covers 12 robot embodiments, including single-arm and bimanual setups, diverse camera views, and a broad range of manipulation behaviors. 
We split the benchmark into 473 validation annotations from AgiBotWorld, DROID, and BridgeData, and 1,275 held-out test annotations. All 438 OXE annotations are test-only and serve as a zero-shot source spanning nine unseen datasets. \gls{method} emits an annotation for 1,256 test samples (424 from OXE), and all methods are evaluated on these same samples.
All reliability-score weights and thresholds were chosen manually on the validation split and frozen before any test evaluation. OXE is never used for tuning.
Unlike static object-grounding benchmarks~\citep{yuan2025fsd, dang2026rynnbrain, robobrain2025, lirobointer}, \gls{bench} includes videos and proprioceptive states, since the manipulated object is often ambiguous in a single frame. This allows evaluating whether a reliability score selects annotations consistent with the actual interaction, rather than ones that are merely plausible in a single image.

\textbf{Evaluation protocol.}
For each demonstration, the pipeline produces candidate annotations consisting of an initial object box, a target object box, and a reliability score $R_i$. 
An annotation is correct, if the bounding box matches ground truth at $\mathrm{IoU}>0.4$ (see threshold ablations in Appendix \ref{app:iou_thresh}).
We evaluate reliability as a selective prediction problem: given all benchmark annotations $\mathcal{D}$, a threshold $\tau$ retains $ \mathcal{D}_\tau = \{i \in \mathcal{D}: R_i \geq \tau\}$.
We report accuracy and coverage to measure the core tradeoff: retaining more data while keeping annotation noise low.

\textbf{Selective reliability metrics.}
We summarize this tradeoff with standard selective-classification metrics \cite{geifman2017selective, feng2018towards, geifman2018bias}. The risk-coverage curve traces annotation error (risk $1 - \text{Acc}(\tau)$) as more annotations are retained, and we report its area (\textbf{AURC}, lower is better) together with the oracle-corrected \textbf{E-AURC}, which isolates the excess risk from imperfect ranking. We further report \textbf{Cov@90}, the largest fraction of annotations retained while keeping retained-set accuracy above $90\%$. Together, these quantify how much usable training data the pipeline can safely produce at a target quality level.

\textbf{Baselines.}
We compare against two baselines and a progressive set of ablations. 
\emph{Det. confidence} ranks candidates based on detector confidence and serves as the strongest static-frame baseline. 
\emph{FSD} \cite{yuan2025fsd} keeps only the highest-confidence detection when its trajectory length exceeds fixed thresholds, and otherwise abstains. Additionally, we compare against external methods that we provide with the demonstration video.

\begin{wrapfigure}{r}{0.6\textwidth}
    \centering
    \vspace{-0.5em}
    \includegraphics[width=\linewidth]{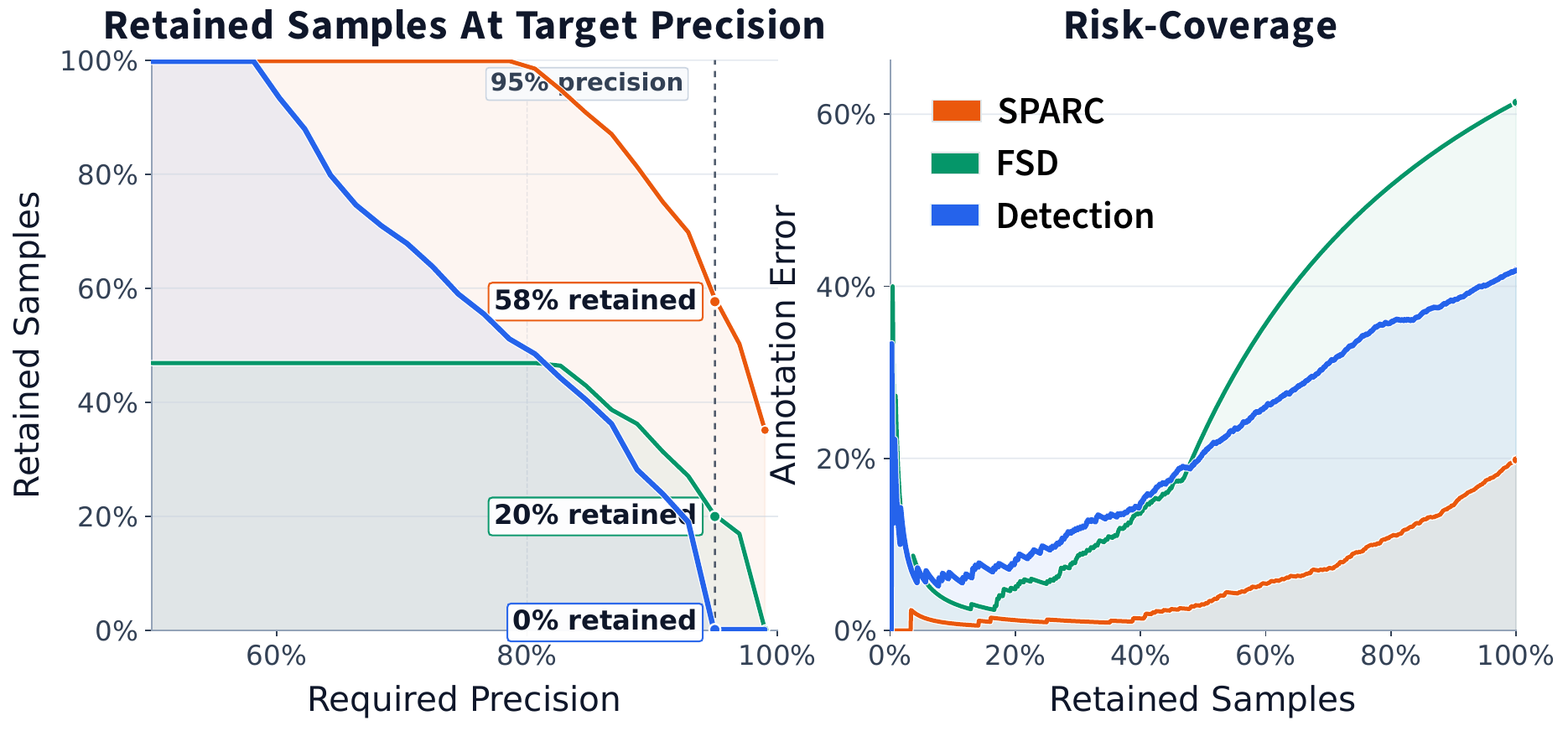}
    \caption{Selective annotation with reliability scoring. 
\emph{Left}: retained sample coverage at increasing target precision thresholds. 
\emph{Right}: risk-coverage curves measuring annotation error as more samples are retained. 
Our reliability score retains more annotations at high target precision and achieves significantly lower risk over all levels, enabling scalable annotation with controllable quality.}
    \label{fig:thresholds}
    \vspace{-1.0em}
\end{wrapfigure}
\textbf{Results.}
Table~\ref{tab:main_results} and Figure~\ref{fig:qualitative_mismatch_comparison} show that prior-style confidence and motion heuristics are insufficient for reliable annotation.
Detector confidence performs moderately across detectors, while raw motion magnitude fails to recover high-precision subsets.
Trajectory length filtering obtains a low E-AURC, but this mainly reflects conservative abstention rather than strong annotation quality.
In contrast, each component of our interaction-aware score improves both selection accuracy and selective ranking (AURC, Cov@90).
Adding object motion gives the largest initial gain, while robot-aware filtering and phase-aware motion further improve accuracy and ranking.
\gls{method} achieves the best results, with \textbf{80.2\% accuracy}, \textbf{77.6\% Cov@90}, and the lowest selective prediction error (\textbf{AURC 0.056}, \textbf{E-AURC 0.035}).
Notably, \gls{method} also exceeds the accuracy of large proprietary embodied models~\cite{team2025gemini} (80.2\% vs.\ 71.7\%), while additionally producing a reliability score for selective annotation.
Figure~\ref{fig:thresholds} further shows that ranking annotations by this score is effective for data filtering.
At a 95\% precision target, \gls{method} retains 58\% of samples, compared to only 20\% for the strongest trajectory-filtering baseline.

\begin{table}[t]
\centering
\footnotesize
\setlength{\tabcolsep}{2pt}
\captionsetup[subtable]{position=top,skip=2pt}
\begin{subtable}[b]{0.565\textwidth}
\centering
\caption{Roles of reliability cues}
\label{tab:cue_roles}
\begin{tabular}{llcccc}
\toprule
& & \multicolumn{2}{c}{Cue alone} & \multicolumn{2}{c}{Removed} \\
\cmidrule(lr){3-4}\cmidrule(lr){5-6}
Cue & Role & Acc. & AUROC & $\Delta$Acc. & $\Delta$Cov@90 \\
\midrule
Motion $A$     & Selector       & .644 & .533 & $-16.9$ & $-42.6$ \\
Detection $D$  & Anchor         & .581 & .774 & $-4.2$  & $-35.4$ \\
Overlap $\Pi$  & Rejector       & .431 & .507 & $-6.8$  & $-13.5$ \\
Proximity $P$  & Conditional    & .473 & .525 & $-0.6$  & $-3.3$ \\
Depth relief   & Conditional    & .291 & .425 & $-0.2$  & $-0.2$ \\
\midrule
Full \gls{method} & --          & .802 & .856 & --      & -- \\
\bottomrule
\end{tabular}
\end{subtable}\hfill
\begin{subtable}[b]{0.42\textwidth}
\centering
\caption{Transfer of the reliability score}
\label{tab:transfer}
\begin{tabular}{lrcccc}
\toprule
& & \multicolumn{2}{c}{Acc.} & \multicolumn{2}{c}{Cov@90} \\
\cmidrule(lr){3-4}\cmidrule(lr){5-6}
Setting & $n$ & Det. & Ours & Det. & Ours \\
\midrule
Complete cycle    & 674 & .588 & \textbf{.800} & .223 & \textbf{.789} \\
Imperfect cycle   & 466 & .571 & \textbf{.820} & .281 & \textbf{.783} \\
No closure        & 116 & .586 & \textbf{.741} & .362 & \textbf{.690} \\
OXE (0-shot)      & 424 & .587 & \textbf{.835} & .380 & \textbf{.863} \\
EgoLive$^\dagger$ & 733 & .492 & \textbf{.754} & .150 & \textbf{.575} \\
\midrule
All test & 1,256 & .581 & \textbf{.802} & .266 & \textbf{.776} \\
\bottomrule
\end{tabular}
\end{subtable}
\caption{\textbf{Reliability score analysis.} (a) Accuracy and AUROC of each cue alone on the IA-Bench test set ($n{=}1{,}256$), and the change in percentage points when removing it from the full score. (b) Detector confidence vs.\ \gls{method}. The first three rows split the test set by detected phase structure. $^\dagger$Egocentric human videos with hand instead of gripper proximity.}
\label{tab:score_analysis}
\vspace{-0.8em}
\end{table}

\subsection{What Makes a Good Reliability Score?}
\label{sec:score_analysis}
A reliability score has two jobs. It must \emph{select} the interacted object among the candidates, and it must \emph{rank} the resulting annotations so that thresholding removes errors first.
Table~\ref{tab:cue_roles} shows that no single cue does both.
Phase-aware motion is the strongest selector (64.4\% accuracy alone) but barely separates correct from incorrect annotations (AUROC .533).
Detector confidence is a weak selector yet the best ranker, since its scale is comparable across demonstrations. Removing it costs only 4.2 accuracy points but 35.4 Cov@90 points.
Robot overlap is weak on its own but supplies negative evidence, rejecting gripper and arm tracks that motion would otherwise favor.
Proximity and depth relief act conditionally. Their average effect is small, but proximity improves ranking when the detector is uncertain, and depth relief matters when the object visually overlaps the gripper (Appendix~\ref{app:cue_analysis}).
Combining the cues therefore helps most when appearance is ambiguous. On the third of samples with the lowest detector margin, accuracy roughly doubles from 35.8\% to 73.5\%.

The cues matter more than their exact weights.
Validation accuracy varies by less than four points across 243 joint weight configurations, and a logistic-regression combiner fit on the same validation data performs worse, particularly on zero-shot OXE (.809 vs.\ .835 accuracy, Appendix~\ref{app:weights}).
Learning the combination thus improves neither accuracy nor transfer.

\subsection{Does the Score Transfer Beyond Grasp Cycles and Robots?}
\label{sec:transfer}
Stage~1 relies on gripper closures to segment interactions. When no closure is observed, e.g., for pushing, \gls{method} falls back to a single interaction phase spanning the subtask.
Table~\ref{tab:transfer} shows that this fallback is clearly weaker than complete grasp cycles (74.1\% vs.\ 80.0\% accuracy) but still well above detector confidence, while imperfect cycles cause no degradation.
Gains also hold for non-prehensile and articulated tasks (46.8\% to 66.1\% accuracy), although press and switch tasks remain hard because static contact produces almost no motion signal (Appendix Table~\ref{tab:per_task}).
On OXE, which was never used for tuning, the fixed score transfers without recalibration.
To test a larger shift, we apply \gls{method} to 733 egocentric human demonstrations from EgoLive~\cite{li2026egolive}, which feature moving head-mounted cameras and hands instead of grippers.
We keep the cue roles but replace gripper proximity with hand proximity from the dataset's hand keypoints, and compensate head rotation with the provided camera poses (Appendix~\ref{app:egolive}).
This raises Cov@90 from .150 to .575 over detector confidence, although less accurate phase estimates and hand masks limit performance.
Thus, what transfers is the role of each cue rather than its robot-specific instantiation.

\subsection{Does Reliability-Based Filtering Improve Embodied Spatial Reasoning?}
\label{sec:downstream}

We next ask whether the reliability score is useful beyond annotation filtering. Concretely, we study whether data generated by \gls{method} provides better supervision for embodied VLM training.

\textbf{Experimental Setup.}
We first curate a VQA dataset spanning target location, vacant pointing, and trace prediction, comprising around 511K VQA pairs. Details of dataset creation are presented in Appendix~\ref{app:vqa_dataset}. Since we use templated language, we co-train with data from LLava-OneVision2~\cite{llava_one_vision_2} and RoboPoint~\cite{yuanrobopoint} to retain general instruction following capabilities. 
We fine-tune Qwen3.5-4B on the same base mixture and add different data mixtures on top to isolate effects coming from the co-training mixture. EO-1.5M~\cite{qu2025eo} is a human-annotated dataset with similar sources and spatial supervision, from which we sample the spatial subsets.

\begin{table*}[t]
\centering
\scriptsize
\setlength{\tabcolsep}{3pt}
\renewcommand{\arraystretch}{1.2}
\resizebox{\textwidth}{!}{%
\begin{tabular}{
l
*{7}{c}
@{\hspace{1.5em}} %
*{5}{c}
@{\hspace{1.5em}} %
*{3}{c}
}
\toprule
&
\multicolumn{7}{c}{\textbf{Pointing} $\uparrow$}
&
\multicolumn{5}{c}{\textbf{Trajectory benchmarks} $\downarrow$} 
&
\multicolumn{3}{c}{\textbf{VQA} $\uparrow$} \\
\cmidrule(lr){2-8}
\cmidrule(lr){9-13}
\cmidrule(l){14-16}
\textbf{Method}
& \makecell{IA \\Bench}
& \makecell{Where2\\Place}
& \makecell{Ref\\Spatial}
& \makecell{RoboSpatial\\Context}
& \makecell{Robo\\RefIt}
& \makecell{VA\\Bench-P}
& \makecell{\textbf{Avg.}}
& \makecell{RoboInter\\Gripper}
& \makecell{RoboInter\\Traj.}
& \makecell{ShareRobot\\Bench-T}
& \makecell{VA\\Bench-V}
& \makecell{\textbf{Avg.}} 
& \makecell{RoboSpatial\\VQA}
& ERQA 
& \makecell{EO\\Bench} \\
\midrule

Qwen3.5-4B
& 49.7      %
& 47.0      %
& 30.5      %
& 35.2      %
& 70.1      %
& 19.0      %
& 41.9      %
& 0.180     %
& 0.282     %
& $\infty$  %
& 0.219     %
& $\infty$  %
& 40.0      %
& 30.3      %
& 30.5 \\   %
\midrule

\textbf{+} Base Mixture
& 45.2      %
& 56.0      %
& 37.3      %
& 44.3      %
& 74.0      %
& 32.3      %
& 48.2      %
& 0.278     %
& 0.298     %
& 0.284     %
& 0.210     %
& 0.268     %
& 65.15     %
& 42.8      %
& 36.3 \\ %

\textbf{+} Detection quality
& 57.7      %
& 53.0      %
& 41.3      %
& 50.0      %
& 84.8      %
& 16.0      %
& 50.5      %
& 0.156     %
& 0.320     %
& 0.242     %
& 0.152 %
& 0.218     %
& 67.4      %
& \textbf{45.6} %
& 37.9 \\   %

\textbf{+} EO-1.5M\cite{qu2025eo}
& 68.9      %
& 70.0      %
& 53.3      %
& 59.0      %
& 85.0      %
& 39.3      %
& 62.6      %
& 0.148     %
& \textbf{0.272} %
& 0.234     %
& $\infty$  %
& $\infty$  %
& \textbf{69.45}     %
& 42.5      %
& \textbf{42.0} \\   %
\midrule
\multicolumn{16}{l}{\textit{Open-source brain models (different base / data). Shown for reference}} \\
\midrule
RoboBrain 2.0 (7B) \cite{team2025robobrain}
& -- & 63.5 & 54.0 & 54.2 & 70.4 & 26.67 & 53.8
& -- & -- & 0.236 & -- & --
& -- & 30.3 & -- \\
MiMo-Embodied (7B) \cite{hao2026mimoembodiedxembodiedfoundationmodel}
& -- & 63.6 & 48.0 & 61.8 & 82.3 & 46.9 & 60.5
& -- & -- & -- & -- & --
& -- & 46.7 & -- \\
\midrule
\rowcolor{gray!15}
\textbf{Ours}
& \textbf{79.1} %
& \textbf{71.0} %
& \textbf{54.5} %
& \textbf{61.0} %
& \textbf{86.4} %
& \textbf{65.7} %
& \textbf{69.6} %
& \textbf{0.125} %
& 0.319     %
& \textbf{0.232} %
& \textbf{0.091} %
& \textbf{0.192} %
& 62.4        %
& 44.9 %
& 35.3 \\   %
\bottomrule
\end{tabular}%
}
\caption{
Grouped downstream benchmark performance.
RefSpatial, RoboSpatial, and RoboRefIt are averaged over their respective subtasks.
Acc. Avg. averages all accuracy-style benchmarks.
Trajectory metrics report RMSE. Bold: best controlled-mixture model.
}
\label{tab:downstream_grouped}
\end{table*}

\textbf{Results.}
Table~\ref{tab:downstream_grouped} demonstrates the effectiveness of \gls{method} for downstream embodied VLM training. Qwen3.5-4B finetuned on \gls{method}-annotated data outperforms all other annotation strategies on diverse spatial pointing tasks, including vacant-space and affordance pointing, and even achieves state-of-the-art results on most benchmarks, as depicted in Table~\ref{tab:downstream_brain_models} in the Appendix. \gls{method} yields stronger grounding than both the human-annotated EO-1.5M data and the noisier detection-annotated data, highlighting the value of reliable auto-annotation and confidence-based filtering.
\gls{method} also provides effective supervision for trajectory prediction, despite training only on object trajectories rather than explicit gripper trajectories. 
On general VQA benchmarks, \gls{method} does not improve over the other mixtures: it trails EO-1.5M on RoboSpatial-VQA (62.4 vs.\ 69.5) and EO-Bench (35.3 vs.\ 42.0), and the detection mixture on ERQA (44.9 vs.\ 45.6).
Our VQA data uses a small set of fixed question and answer templates to isolate the effect of annotation quality, whereas EO-1.5M relies on human-refined and LLM-rewritten prompts that better match open-ended questions.
More diverse VQA pairs can be generated cheaply from the same annotations.

\begin{figure}[t]
    \centering
    \includegraphics[width=\linewidth]{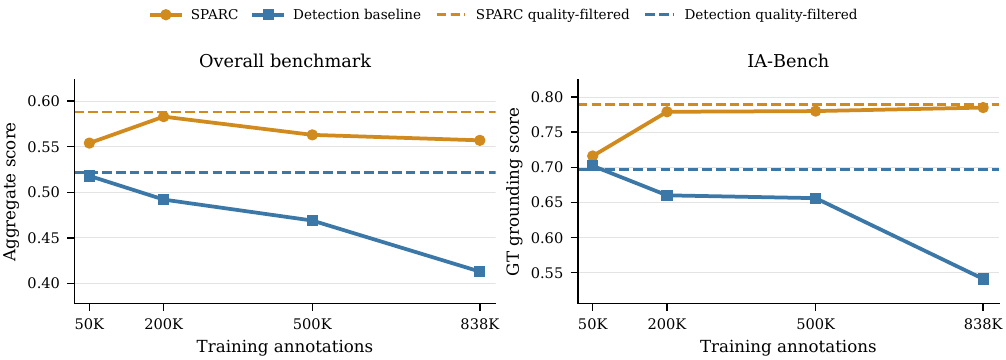}
\caption{\textbf{Scaling behavior across downstream benchmarks.}
We train on the top-scoring 50K, 200K, 500K, and 838K annotations selected by SPARC or detector confidence. Dashed lines show the quality-filtered setting using the fixed threshold from the main experiments. Left: average performance across all downstream benchmarks. Right: performance on IA-Bench.}
\label{fig:sparc_scaling_combined}
\end{figure}

The reliability score also matters when scaling the training data (Figure~\ref{fig:sparc_scaling_combined}).
When training on the top-ranked 50K to 838K annotations, models trained on detector-selected data degrade as more samples are added.
In contrast, \gls{method} keeps improving on \gls{bench} (from .716 to .785) and peaks at 200K samples on the full benchmark suite (Appendix~\ref{app:scaling}).

\subsection{Do High-Quality Spatial Annotations Improve Real-World Policy Performance?}
\begin{wraptable}{r}{0.46\textwidth}
\centering
\footnotesize
\setlength{\tabcolsep}{2pt}
\begin{tabular}{lcc}
\toprule
Policy & Grounding & Success \\
\midrule
Baseline               & 21 & 9 \,[4.8, 16.2] \\
Detection Reasoning    & 21 & 12 \,[7.0, 19.8] \\
\gls{method} Reasoning & \textbf{64} & \textbf{31} \,[22.8, 40.6] \\
\bottomrule
\end{tabular}
\caption{Real-world results out of 100 rollouts (10 tasks $\times$ 10). Grounding: rollouts in which the policy reaches for the correct object. Brackets: 95\% Wilson confidence intervals.}
\label{tab:policy_performance}
\end{wraptable}

We construct a cluttered tabletop setting where the robot must move visually similar objects to target locations, so that success depends almost entirely on correct grounding rather than on low-level control. We train three VLAs on 250 demonstrations across 10 tasks, all using a Qwen3.5-0.8B backbone with a flow-matching action head~\cite{ye2026starvla}, following recent VLA designs~\cite{reuss2025flower, zhou2025beast, vanjani2026damvla}. The baseline trains only on next action prediction. We further co-train two reasoning policies that predict trajectories as language before actions, with data annotated by the detection baseline and \gls{method}, denoted Detection Reasoning and \gls{method} Reasoning. As shown in Table~\ref{tab:policy_performance}, \gls{method} Reasoning more than doubles the success rate of
the detection-annotated policy and triples that of the no-reasoning baseline, under identical training conditions, and both improvements are statistically significant.
\gls{method} Reasoning matches or exceeds both policies on every task and is strictly better on seven of ten (Appendix Table~\ref{tab:real_robot_per_task_counts}), suggesting that the quality of the reasoning annotations, rather than policy capacity, drives the gain on these grounding-heavy tasks.
It also grounds the correct object three times as often as both other policies, consistent with grounding being the source of the gain.

\section{Conclusion}
\label{sec:conclusion}

In this work, we introduce \gls{method}, which treats automatic spatial annotation of robot demonstrations as selective prediction and scores each annotation with physical interaction evidence rather than detector confidence.
The resulting reliability score makes the trade-off between quality and scale controllable without human review.
Our analysis shows that interaction cues play complementary roles, with motion selecting the manipulated object, detector confidence anchoring the ranking, and robot overlap rejecting the embodiment, and that these roles transfer to unseen embodiments and egocentric human videos.
Models trained on \gls{method} annotations surpass those trained on human-annotated data on object-grounding and pointing benchmarks, and reasoning policies trained on them are substantially more robust in cluttered real-world manipulation.

\textbf{Limitations.} \gls{method} is bounded by the off-the-shelf models it builds on, such as detector errors on object parts, tracker identity switches between similar objects, and 3D lifting of transparent surfaces.
Subtask decomposition relies on gripper closures, and phase errors propagate downstream. Without closures, \gls{method} falls back to a single interaction phase, which is clearly weaker, and visually static contact, as in pressing buttons, yields almost no motion signal.
Contact events from force-torque or tactile sensing could replace visual grasp phases in such tasks.
\gls{method} also assumes a mostly static camera. EgoLive shows that known camera poses can compensate head motion, but strong unknown camera motion remains open.
Finally, whether the gains hold for weaker or stronger perception models remains to be tested.

\clearpage
\acknowledgments{This work was funded by the Deutsche Forschungsgemeinschaft (DFG, German Research Foundation) -- 448648559. The authors gratefully acknowledge the computing time provided on the high-performance computer HoreKa by the National High-Performance Computing Center at KIT (NHR@KIT). This center is jointly supported by the Federal Ministry of Education and Research and the Ministry of Science, Research and the Arts of Baden-W\"urttemberg, as part of the National High-Performance Computing (NHR) joint funding program (\url{https://www.nhr-verein.de/en/our-partners}). HoreKa is partly funded by the German Research Foundation (DFG). The authors gratefully acknowledge the Gauss Centre for Supercomputing e.V. (\url{www.gauss-centre.eu}) for supporting this project by providing computing time on the GCS Supercomputer JUPITER at J\"ulich Supercomputing Centre (JSC). The authors gratefully acknowledge the support of the Robotics Institute Germany (RIG).}

\bibliography{references}

\newpage
\appendix
\label{app:ablation}

\section{Additional Related Work}
\label{app:related_work}

\begin{table*}[h]
\centering
\small
\setlength{\tabcolsep}{4.5pt}
\renewcommand{\arraystretch}{1.15}
\resizebox{\linewidth}{!}{%
\begin{tabular}{l|l|c|C{0.85cm}C{0.85cm}C{0.85cm}|l}
\toprule
\textbf{Type} &
\textbf{Method / dataset} &
\textbf{Ann.} &
\multicolumn{3}{c|}{\textbf{Filtering signal}} &
\textbf{Labels} \\
& & &
\textbf{Det.} & \textbf{Track} & \textbf{Robot} &
\\
\midrule
\multirow{3}{*}{\rotatebox{90}{\textit{Human}}}
& RoboInter~\cite{lirobointer} / RynnBrain~\cite{dang2026rynnbrain} & H & \xmark & \xmark & \xmark & Box / VQA / contact \\
& EO-1M~\cite{qu2025eo} & H+A & \xmark & \xmark & \xmark & Box / VQA / contact \\
& AgiBot-World~\cite{agibot2025} / VeBrain~\cite{luo2025visual} & H & \xmark & \xmark & \xmark & Task / scene / object \\
& Gemini Robotics~\cite{team2025gemini} & ? & ? & ? & ? & Trace / Mask\\
\midrule
\multirow{2}{*}{\rotatebox{90}{\textit{Sim.}}}
& RoboPoint~\cite{yuanrobopoint} / InterVLA-A1~\cite{chen2025internvla} & A & \cmark & \xmark & \xmark & Box / CoT \\
& RoboRefer~\cite{zhou2026roborefer} / GraspVLA~\cite{denggraspvla} & A & \cmark & \xmark & \xmark & Box / CoT \\
\midrule
\multirow{7}{*}{\rotatebox{90}{\textit{Automatic filtering}}}
& LaRA-VLA~\cite{bai2026latent} / ECoT~\cite{zawalski2024ecot,chen2025training} & A & \cmark & \xmark & \xmark & Box / CoT \\
& Fast-Think-Act / Molmo-Act~\cite{huang2026fast} & A & \cmark & \xmark & \xmark & Box / CoT \\
& PixelVLA~\cite{liang2024pixelvla} & A & \cmark & \xmark & \xmark & Box / mask \\
& SAM2Auto~\cite{yang2025sam2auto} / FoundationMotion~\cite{gan2025foundationmotion} /FSD~\cite{yuan2025fsd} & A & \cmark & \cmark & \xmark & Track / motion \\
& HALO~\cite{Shou2026HALOAU} & A & \xmark & \xmark & \xmark & CoT / affordance \\
& Robo2VLM~\cite{chen2026robovlm} & A & \xmark & \xmark & \cmark & Grip. phases \\
\midrule
\rowcolor{gray!12}
& \textbf{Ours} & \textbf{A} & \cmark & \cmark & \cmark & \textbf{Box / Mask / Grip / Trace} \\
\bottomrule
\end{tabular}}
\caption{
Comparison of annotation pipelines by annotation source and filtering signal.
\textbf{Type} groups methods by annotations type:
\textit{Human} (manually labelled real-world data),
\textit{Sim.} Simulation-based, and
\textit{Automatic filtering}.
\textbf{Ann.}: H = human, A = automatic.
Det.\ denotes detector or mask confidence.
Ours is the only pipeline combining all three filtering signals with the richest label set.
}
\label{tab:annotation_filtering}
\end{table*}

\Cref{tab:annotation_filtering} shows existing works and how they generate and filter spatial annotations from robot demonstrations. Notably, almost all of them use a detection model as main confidence estimate, while some additionally use tracking as a signal. However, as discussed previously, the initial detector confidence is not calibrated well, thus relying on its confidence followed by track length filtering results in a significantly reduced coverage. On the other hand, \gls{method} combines spatio-temporal properties of robot object manipulation to obtain a robust and well calibrated filtering signal.

\section{Implementation Details and Choice of Foundation Models}

\subsection{\gls{method} Implementation Details}
\label{app:Implementation Details}
\subsubsection{Stage 1 - Subtask Decomposition and Phase Detection}

\textbf{Interaction segmentation.}
Given a long-horizon demonstration
\(
\tau=\{(x_t,q_t)\}_{t=1}^{T},
\)
with RGB frames \(x_t\), proprioceptive state \(q_t\), and language instruction \(\ell\), \gls{method} decomposes \(\tau\) into a sequence of short object-centric subtasks. Each subtask is intended to isolate one coherent interaction with a single manipulated entity, so that later proposal, tracking, and scoring operate on temporally localized clips rather than the full demonstration. For tool-use behaviors, the manipulated tool is treated as the primary interaction object, while the acted-on receptacle or surface is retained as the target context. The output of this stage is a structured subtask set
\[
\mathcal{S}(\tau)
=
\left\{
\left(
u_j,\ell_j,q^{\mathrm{init}}_j,q^{\mathrm{target}}_j,\mathcal{P}_j
\right)
\right\}_{j=1}^{M},
\]
where \(u_j\) is the subtask instruction, \(\ell_j\) the manipulated object name, \(q^{\mathrm{init}}_j\) and \(q^{\mathrm{target}}_j\) textual descriptions of the initial and target locations, and \(\mathcal{P}_j\) the associated grasp-phase sequence.

\textbf{Gripper phase extraction.}
\gls{method} derives the temporal support of each interaction from the normalized gripper signal \(a_t\in[0,1]\), where smaller values correspond to a more closed gripper. The implementation uses a \emph{dual-pass} procedure with hysteresis to robustly separate true grasps from teleoperation noise, short slips, and transient re-openings. We first normalize the raw gripper trace per trajectory and define asymmetric closing and opening thresholds
\[
\tau_{\mathrm{enter}}=\tau_c-\delta,
\qquad
\tau_{\mathrm{exit}}=\tau_c+\delta,
\]
so the state enters a closed interaction only when \(a_t<\tau_{\mathrm{enter}}\) and exits it only when \(a_t\ge\tau_{\mathrm{exit}}\). This hysteresis prevents oscillatory threshold crossings from fragmenting a single interaction into many short phases. If the gripper signal shows almost no range, the trajectory is treated as a single coarse interaction span.

In the \emph{first pass}, \gls{method} scans the trajectory with a finite-state machine over \texttt{grasp}, \texttt{interact}, and \texttt{release}. A transition is only accepted after several consecutive frames satisfy the new state condition, which suppresses one-frame fluctuations. This pass does \emph{not} yet decide whether a closure is valid or failed; instead, it simply records candidate closed-gripper spans and their durations
\[
d_k = e_k-s_k+1,
\]
where \([s_k,e_k]\) is the \(k\)-th candidate interaction interval. From these provisional spans, the algorithm estimates a robust per-trajectory interaction scale. Concretely, it computes the median candidate duration and combines it with a per-attempt interaction budget derived from the trajectory length and the number of detected attempts, yielding a minimum valid interaction duration
\[
d_{\min}
=
\max\!\left(
0.5\,\frac{B}{K},
\,0.4\,\mathrm{median}(\{d_k\}_{k=1}^{K})
\right),
\]
where \(B\) is the expected total interaction budget for the trajectory and \(K\) is the number of candidate closures. This makes the threshold adaptive: in short or repetitive demonstrations, valid short grasps are preserved, while unusually brief closures remain identifiable as outliers.

In the \emph{second pass}, the same state machine is run again, now using \(d_{\min}\) to classify each closed interval. Closures whose interact duration exceeds \(d_{\min}\) are emitted as standard \texttt{grasp} \(\rightarrow\) \texttt{interact} \(\rightarrow\) \texttt{release} cycles, whereas shorter closures are relabeled as \texttt{grasp\_failure} and do not create full interaction segments. The implementation additionally repairs boundary artifacts in two ways. First, if a new grasp begins too soon after a release, a small number of frames may be reassigned from the preceding release to the new grasp so that the new cycle retains a plausible approach phase. Second, release phases are allowed to terminate early when the gripper re-closes quickly, ensuring back-to-back pick attempts are not merged. After phase extraction, \gls{method} applies a small forward offset to each phase boundary, borrowing a fraction of the following phase, so that downstream keyframes better capture contact transitions and early object motion. For bimanual trajectories, this entire dual-pass procedure is applied independently to each arm before the two phase streams are merged temporally.

\textbf{Language-grounded subtask parsing.}
Gripper phases indicate when interaction occurs, but not which object is involved or how the interaction should be described. \gls{method} therefore conditions a language model on the original instruction \(\ell\) together with the ordered phase list and asks it to recover a structured description of the executed subtasks. For each subtask, the model returns the manipulated object name, action type, start and target locations, and a phase-level natural-language description aligned with the detected phase sequence. This language grounding step is cached by instruction together with the detected phase-sequence key, so repeated trajectories with the same language-phase pattern do not trigger repeated model queries. In bimanual clips, the prompt is arm-aware and predicts arm-specific object assignments before the two streams are merged back into a unified subtask list.

\textbf{Subtask boundary alignment.}
After parsing, the symbolic phase descriptions are aligned back to the measured trajectory timeline. Concretely, each predicted phase entry inherits its start and end frames from the ordered detected phase list, producing a temporally grounded subtask clip for downstream annotation. \gls{method} then expands each subtask clip slightly beyond the raw closed-gripper interval using the offset phases, so proposal and tracking can observe approach motion before contact and release motion after placement. The result is a set of short-horizon interaction records that jointly specify \emph{what} object to look for, \emph{when} to look for it, and \emph{where} it is expected to begin and end.

\subsubsection{Stage 2 - Candidate Proposal, Segmentation, and Tracking}

\textbf{Object proposal at interaction keyframes.}
For each subtask \(j\), \gls{method} extracts the corresponding video clip and selects a grasp-conditioned detection keyframe near the onset of object interaction. The manipulated object is then proposed with LLMDet~\cite{fu2025llmdet} using a low confidence threshold, so the detector acts primarily as a broad region proposal mechanism rather than a strict final classifier. To calculate the grounding scores, we average the token logits for each box.
In parallel, RobotSeg segments the robot arm and gripper at the same keyframe, which later supports candidate filtering, robot-aware scoring, and failure recovery. This proposal stage intentionally favors recall: it is preferable to include several plausible object boxes and defer disambiguation to later tracking and scoring.

\textbf{Target proposal.}
The final object location is not inferred from motion alone. Instead, \gls{method} also proposes target boxes using the target-location phrase returned in Stage~1. The implementation queries target candidates from early and late frames of the subtask clip, then merges and suppresses them with size filtering and non-maximum suppression. For pick-and-place-like behaviors, this gives a set of plausible end locations against which tracked object motion can be matched. If the language model did not predict an explicit target location, the pipeline falls back to the object or tool name as a weak target descriptor and detects on the last frame rather than discarding the subtask.

\textbf{Segmentation and candidate refinement.}
Each detected object proposal is refined into an instance mask with SAM2~\cite{ravi2025sam}, yielding a set of candidates
\[
c_i = \left(b_i,m_i,\ell_i,D_i\right),
\]
where \(b_i\) is the initial box, \(m_i\) the segmentation mask, \(\ell_i\) the text-grounded label, and \(D_i\) the detector confidence. Mask refinement is important because later tracking operates on object-support regions rather than on coarse boxes alone. For DROID-like data, the implementation may crop the clip around the union of detected object and target boxes before tracking, which reduces wasted image area and improves tracking stability in wide views; all coordinates are restored to the original image frame before the final annotation is written.

\textbf{Temporal tracking and 3D lifting.}
Given the segmented candidates, \gls{method} propagates object evidence over time and ranks candidates by whether their trajectories match the intended interaction. The current implementation supports several tracking variants: a point-based tracker, a SAM2-video propagation mode, and a hybrid mode in which an initial point-tracker ranking is re-evaluated with SAM2 on the top candidates. At a high level, all variants produce temporally extended candidate representations of the form
\[
o_i
=
\left(
c_i,\Gamma_i^{2\mathrm{D}},\Gamma_i^{3\mathrm{D}}
\right),
\]
where \(\Gamma_i^{2\mathrm{D}}\) denotes the candidate's propagated image-space tracks or masks across the subtask clip, and \(\Gamma_i^{3\mathrm{D}}\) denotes the corresponding lifted 3D tracks when geometry is available. The 3D lifting step is optional in the sense that it is only applied when dense geometry prediction succeeds, but when available it provides embodiment-aware evidence for later scoring. Intermediate detections at additional keyframes are also retained, allowing the tracker to prefer candidates whose propagated support remains compatible with re-detections throughout the clip rather than only at the start and end.

\subsubsection{Stage 3 - Reliability Scoring and Final Annotation Selection}

Stage~2 yields a set of tracked object candidates
\(
\{o_i\}_{i=1}^{N}
\),
where each candidate consists of an initial detection, a segmentation mask, and a set of tracked object points over time. Stage~3 ranks these candidates by combining appearance reliability, phase-aware motion, 3D gripper proximity, and a robot-overlap penalty. The score is designed so that the selected object is not merely visually plausible, but also exhibits the temporal and spatial signatures expected of a manipulated object.

\textbf{Per-candidate tracked support.}
For each candidate \(o_i\), tracking starts from points sampled inside the first-frame object mask \(m_i\). Let
\(
\Gamma_i=\{\mathbf{u}_{i,p}^{t}\}
\)
denote the resulting 2D tracks, where \(p\) indexes sampled points and \(t\) indexes frames in the subtask clip. Each point also has a visibility indicator \(v_{i,p}^{t}\in\{0,1\}\). Only points that remain visible for a sufficient fraction of the clip are retained for scoring, which suppresses unstable or spuriously short tracks. All motion, target, and proximity signals are then computed from this filtered point set.

\textbf{Detector confidence.}
Each candidate \(o_i\) inherits a language-grounded detector confidence \(D_i\) from the initial proposal stage. This term captures appearance-level compatibility between the visual region and the queried object description. On its own, \(D_i\) is often insufficient for selecting the manipulated object, since visually salient distractors or robot parts can also receive high scores. It is therefore used as an appearance prior that is later reweighted by geometric interaction evidence rather than as a standalone selection rule.

\textbf{Phase-aware motion signal.}
The first signal measures whether a candidate moves primarily during the interaction phase rather than throughout the full clip. For each visible point, we compute frame-to-frame displacement and then aggregate it at the candidate level. Concretely, for candidate \(i\), the instantaneous 2D motion at frame transition \(t\rightarrow t+1\) is
\[
m_{i,t}
=
\operatorname{median}_{p:\,v_{i,p}^{t}=v_{i,p}^{t+1}=1}
\left\|
\mathbf{u}_{i,p}^{t+1}-\mathbf{u}_{i,p}^{t}
\right\|_2
\cdot \mathrm{fps},
\]
that is, the median displacement of all visible tracked points, converted to pixels per second. A short temporal smoothing window is then applied to reduce jitter. Let \(\mathcal{T}_{\mathrm{int}}\) denote the set of interaction-phase frame transitions and \(\mathcal{T}_{\mathrm{non}}\) the remaining valid transitions. Their average motion is
\[
\mu_i^{\mathrm{int}}
=
\frac{1}{|\mathcal{T}_{\mathrm{int}}|}
\sum_{t\in\mathcal{T}_{\mathrm{int}}} m_{i,t},
\qquad
\mu_i^{\mathrm{non}}
=
\frac{1}{|\mathcal{T}_{\mathrm{non}}|}
\sum_{t\in\mathcal{T}_{\mathrm{non}}} m_{i,t}.
\]
The raw phase-aware motion score is then computed as a soft signal-to-noise ratio,
\[
A_i
=
\frac{
(\mu_i^{\mathrm{int}})^{0.6}
}{
(\mu_i^{\mathrm{non}}+1.0)^{0.2}
}.
\]
This favors candidates that move persistently when the gripper is engaged, but penalizes candidates whose motion is equally strong before grasp or after release, which is typical of robot parts, background clutter, or unstable tracks. The set \(\{A_i\}\) is min-max normalized across candidates in the same clip to obtain \(\widehat{A}_i\).

\textbf{3D gripper-proximity signal.}
The second signal measures whether the candidate occupies the local 3D neighborhood of the active gripper during manipulation. When dense geometry is available, each tracked 2D point is lifted to 3D, giving per-candidate 3D tracks
\(
\mathbf{x}_{i,p}^{t}\in\mathbb{R}^{3}.
\)
A gripper reference point is then estimated for each frame from the tracked gripper points. Rather than averaging all gripper points, which is unstable under partial visibility and self-occlusion, the reference is chosen as the gripper point whose 2D projection lies closest to the image center among the visible gripper points in that frame. This choice produces a stable central proxy for the fingertip/TCP region. We found that selecting a consistent track is unreliable becuase trackers struggle significantly with tracking a point on the gripper consistently, likely due to visual similarity accross the gripper.

Around this reference point \(g_t\), an adaptive 3D sphere \(B(g_t,r_t)\) is constructed. Its radius is determined from the local gripper geometry together with an estimate of object scale, so the same criterion remains meaningful across small and large objects. \(Q_{i,t}\) is the set of visible 3D points of candidate \(i\) at frame \(t\). The per-frame proximity fraction is
\[
p_{i,t}
=
\frac{1}{|Q_{i,t}|}
\sum_{\mathbf{x}\in Q_{i,t}}
\mathbf{1}\!\left[\mathbf{x}\in B(g_t,r_t)\right].
\]
Averaging over valid frames yields the 3D proximity score
\[
P_i
=
\frac{1}{|\mathcal{T}_i|}
\sum_{t\in\mathcal{T}_i} p_{i,t}.
\]
In the final implementation, the mean point-fraction variant is used, and the sphere radius is slightly enlarged relative to the raw gripper scale so that nearby in-contact objects are not undercounted. The resulting \(\{P_i\}\) are min-max normalized across candidates to obtain \(\widehat{P}_i\).

\textbf{Robot-overlap penalty.}
A major failure mode is selecting the gripper or robot arm itself. This is addressed with a continuous penalty rather than a hard rejection rule. For bimanual data, the RobotSeg mask is restricted to the active arm component before scoring, which prevents the inactive arm from suppressing candidates on the correct side of the workspace. Active arms are determined by activity in the proprio gripper signal. 

At the detection keyframe, the candidate's visible tracked points are rasterized against the RobotSeg mask, yielding a candidate-level overlap score \(O_i\in[0,1]\). High overlap should decrease the final score, but not all overlap is equally harmful: true objects frequently touch or partially occlude the gripper. To distinguish contact from identity confusion, the penalty is modulated by a 3D depth-relief term. Specifically, the median absolute depth difference between the gripper tip region and the candidate tip region is measured across valid frames. Large depth separation weakens the overlap penalty, while near-zero separation preserves it.

The final robot penalty \(\Pi_i\) is therefore a continuous quadratic function of RobotSeg overlap, attenuated by this depth-relief factor. This is crucial in practice: earlier hard overlap filtering removed many correct contact-rich candidates, whereas the continuous penalty suppresses obvious robot regions without catastrophically rejecting true manipulated objects.

\textbf{Composite score.}
The final reliability score follows the notation used in the main paper. We min-max normalize \(A_i\) and \(P_i\) across candidates in the same clip, giving \(\widehat{A}_i\) and \(\widehat{P}_i\), and combine all four signals:
\[
R_i
=
w_A\,\widehat{A}_i + w_D(\widehat{P}_i)\,D_i + w_P\,\widehat{P}_i - \Pi_i,
\]
where \(w_A=0.5\), \(w_P=0.3\), and
\[
w_D(\widehat{P}_i)=0.75-0.15\,\widehat{P}_i.
\]
Thus, detector confidence \(D_i\) contributes strongly when 3D proximity evidence is weak, but its influence is reduced once the candidate is already well supported by embodiment-aware context. The selected manipulated object is
\[
i^\star=\arg\max_i R_i.
\]

\textbf{Final target selection and target position estimation.}
The selected manipulated object is
\(
i^\star=\arg\max_i R_i.
\)
After selecting \(i^\star\), \gls{method} resolves the target position from a set of detected target candidates
\(
\mathcal{C}^{\mathrm{tar}}=\{c^{\mathrm{tar}}_j\}_{j=1}^{M},
\)
obtained from language-conditioned target detection on early and late frames of the subtask clip. In the main pipeline, these candidates are scored jointly with the selected object hypothesis using motion, temporal alignment, and target agreement, so target resolution remains tied to the same interaction evidence used for manipulated-object selection.

Concretely, \gls{method} measures how many tracked points from the selected object candidate terminate inside each target candidate and prefers targets that receive a dense concentration of transported points. In downstream target remapping, this support is normalized by target-box size to avoid favoring large diffuse boxes. Let \(S_{i^\star j}^{\mathrm{ungated}}\) denote the fraction of visible final tracked points from candidate \(i^\star\) that lie inside target candidate \(c_j^{\mathrm{tar}}\), and let \(A_j\) denote its box area. The density-normalized target score is
\[
\widetilde{S}_{i^\star j}
=
\frac{
S_{i^\star j}^{\mathrm{ungated}}
}{
\sqrt{A_j/A_{\max}}
},
\qquad
A_{\max}=\max_j A_j.
\]
This favors compact target boxes that capture a high density of transported points. To avoid degenerate tiny boxes, candidates whose area is less than half the selected start-box area are excluded before ranking.

If tracking-supported target resolution is unavailable, \gls{method} falls back to the best detected target candidate; for tool-use tasks, it further falls back to the acted-on object or tool context when no separate target box is available. Note that since the best interacted object is already fixed, which includes movement scoring, this approach does not degenerate to static points, which would dominate this selection logic.

\subsection{Runtime}
\label{app:runtime}
\gls{method} runs several large foundation models in sequence for each trajectory, resulting in roughly 5 seconds per annotation on a single NVIDIA GH200 GPU.
To reduce this further, we subsample tracking frames to 6\,fps. We keep the native resolution, since lowering it degraded annotation quality, especially for small objects.
For reference, human annotators labeled the 1{,}275 trajectories in the \gls{bench} test set in roughly 12 hours, corresponding to approximately 34 seconds per annotation.
Crucially, \gls{method} is highly parallelizable, as trajectories can be processed independently.
To maximize GPU utilization, each GPU hosts a dedicated inference server that loads all models once at startup and exposes them to multiple CPU workers through an asynchronous request queue.
Each CPU worker processes a separate trajectory and issues inference requests concurrently, keeping the GPU continuously occupied across detection, segmentation, and tracking.
On a single node with four GH200 GPUs and four CPU workers per GPU, 16 workers in total, annotating the same 1{,}275 trajectories takes approximately 30 minutes.
This yields a $24\times$ wall-clock speedup over human annotation, and the workload distributes across multiple nodes without modification. Furthermore, all annotations generated by our pipeline can be used downstream for several applications.

\subsection{Foundation Model Choices}
\label{app:foundation_model_choices}

\begin{table}[H]
\centering
\small
\setlength{\tabcolsep}{4pt}
\renewcommand{\arraystretch}{1.12}
\begin{tabular}{>{\raggedright\arraybackslash}p{0.13\linewidth} >{\raggedright\arraybackslash}p{0.21\linewidth} >{\raggedright\arraybackslash}p{0.27\linewidth} >{\raggedright\arraybackslash}p{0.29\linewidth}}
\toprule
\textbf{Component} & \textbf{Model} & \textbf{Input} & \textbf{Output} \\
\midrule
Task parsing & Qwen3.6-30B-MOE~\cite{qwen35blog} & Task instruction and detected gripper phases & One JSON record per subtask with object, start and target locations, action, tool use, and phase descriptions \\
Object detection & LLMDet-Base~\cite{fu2025llmdet} & Detection keyframe and the parsed object or target name as text prompt & Candidate boxes with grounding confidence \\
Segmentation & SAM2.1 Hiera-Small~\cite{ravi2025sam} & Detection keyframe and candidate boxes & Instance mask per candidate \\
Robot segmentation & RobotSeg~\cite{mei2025robotseg} & Detection keyframe & Gripper and arm mask \\
Object tracking & AllTracker~\cite{harley2025alltracker} & Subtask video clip & Dense 2D point tracks with visibility \\
Depth / geometry & MoGe-2~\cite{wang2026moge} & Single RGB frame, optionally camera intrinsics & Per-pixel 3D point map \\
\bottomrule
\end{tabular}
\caption{Foundation models used in \gls{method}, with their inputs and outputs.}
\label{tab:foundation_models}
\end{table}
\gls{method} relies on off-the-shelf foundation models to generate interaction-aware signals used for annotation and reliability scoring. We will explain the choice for each individual model in this section. Note that \gls{method} allows changing the individual models as better models for embodied settings emerge, and the method is not tied to a specific set of models. \Cref{tab:foundation_models} lists all the foundation models used for generating the different candidate representations. For task decomposition in stage 1 \gls{method} leverages Qwen3.6 due to fast inference speed and good out-of-the-box embodied and physical reasoning capabilities. 
For intial candidate proposal with open-vocabulary object detection, we rely on LLMDet. Although OWLv2 is slightly better in grounding objects in robotic settings, we found that its grounding score is generally less calibrated compared to LLMDet. Similar behavior was observed with GroundingDino \cite{liu2023groundingdino}. Furthermore, LLMDet produces less noisy boxes at very low thresholds. Since \gls{method} uses the initial detector mainly as a region proposal network, the proposal network has to be robust at low threshold regimes, and should actually output boxes around objects and not empty space.

For gripper detection, we use RobotSeg, a model that builds on SAM2 and is trained specifically to segment gripper masks in video, producing far more reliable gripper masks than a standard pipeline that follows object detection and segmentation. Existing object detectors and segmentation models struggle to consistently localize and segment the same parts of the gripper.

For object tracking, we rely on a dense tracker instead of sparse trackers such as \cite{karaev2023cotracker}. We found that the tracks are more robust than those from sparse tracking. Furthermore, since we track many possible candidate objects, CoTracker runtime was significantly larger than AllTracker, which despite tracking all pixels in the scene has proven to be very efficient.

\gls{method} uses SAM2.1 for segmentation instead of the newer SAM3 \cite{carion2025sam} since SAM2.1 is significantly faster than SAM3. Furthermore, \gls{method} does not require the advantages coming with SAM3, such as grounding or better masks.

Finally, we rely on a monocular depth prediction model MoGe-2 for 3D lifting. MoGe-2 predicts scene geometry individually per frame from a single RGB observation and outputs a depth map, normals, and point cloud without requiring camera intrinsics. If intrinsics are available, MoGe-2 can use those to improve geometry reconstruction. Because MoGe does not incorporate temporal information, the resulting depth maps differ slightly in scale across frames but remain consistent within each frame. Since the signals used by \gls{method} operate on a per-frame basis, we do not require consistent depth maps on a temporal axis. While there are methods specifically tailored at predicting consistent depth across time \cite{chen2025video}, they often require known intrinsics or have significantly higher runtime. Furthermore, MoGe operates at a higher native resolution, which is important for accurately capturing gripper-object boundaries necessary for robust gripper overlap calculation.

\subsection{Pipeline Hyperparameters and Thresholds}
In \Cref{tab:hyperparams}, we show the hyperparameters used in \gls{method} to perform all evaluations. Note that we did tune some of these hyperparameters on the hold-out validation set, but never on the test set. Furthermore, we did not tune the hyperparameters on the OXE dataset. These parameters have proven robust when transferring to the test set and new embodiments.
\begin{table}[h]
\centering
\small
\begin{tabular}{llc}
\toprule
Name & Symbol & Value \\
\midrule
\multicolumn{3}{l}{\textit{Stage 1 -- Gripper phase detection}} \\
Closure threshold        & $\tau_c$               & 0.6 \\
Hysteresis offset        & $\delta$               & 0.05 \\
Min.\ phase duration     & $T_{\min}$             & 7 frames \\
Phase start offset       & $r_{\text{off}}$       & 0.3 \\
\midrule
\multicolumn{3}{l}{\textit{Stage 2 -- Detection keyframe}} \\
Keyframe position        & $t^*$                  & midpoint of subtask start and first grasp end \\
Detector Threshold & -- & 0.05 \\
\midrule
\multicolumn{3}{l}{\textit{Stage 3 -- Phase-aware motion signal}} \\
Interact exponent        & $\alpha$               & 0.6 \\
Non-interact exponent    & $\beta$                & 0.2 \\
Noise floor              & $\varepsilon$          & 1.0\,px/s \\
Motion term weight       & $w_A$                  & 0.5 \\
\midrule
\multicolumn{3}{l}{\textit{Stage 3 -- Adaptive detector and sphere proximity}} \\
Detector base weight     & $w_D^{\,0}$            & 0.75 \\
Adaptive detector slope  & $w_D^{\,\text{slope}}$ & 0.15 \\
Sphere proximity weight  & $w_P$                  & 0.3 \\
Sphere radius scale      & $k_r$                  & 8 \\
\midrule
\multicolumn{3}{l}{\textit{Stage 3 -- Robot-overlap soft gate}} \\
Overlap onset            & $\tau_{\text{ov}}$     & 0.3 \\
Penalty scale            & $\lambda$              & 1.45 \\
Depth attenuation        & $\gamma$               & 0.85 \\
Full-overlap spike coeff.& $\lambda_{\text{sp}}$  & 0.20 \\
Full-overlap spike thresh.& $\tau_{\text{sp}}$    & 0.98 \\
Depth relief lower bound & $z_\ell$               & 0.05\,m \\
Depth relief upper bound & $z_h$                  & 0.25\,m \\
\midrule
\multicolumn{3}{l}{\textit{Evaluation}} \\
IoU match threshold      & $\tau_{\text{IoU}}$    & 0.4 \\
Containment fallback IoU & $\tau_{\text{ct}}$     & 0.1 \\
Containment fraction     & $f_{\text{ct}}$        & 0.8 \\
\bottomrule
\end{tabular}
\caption{SPARC pipeline hyperparameters. All values are fixed across
datasets and embodiments.}
\label{tab:hyperparams}
\end{table}

\section{Additional Experiments and Analysis}

\subsection{Reliability Score Sensitivity Analysis}
\begin{table}[H]
\centering
\small
\setlength{\tabcolsep}{6pt}
\begin{tabular}{llccc}
\toprule
Parameter & Value & Acc~$\uparrow$ & Cov@90~$\uparrow$ & AURC~$\downarrow$ \\
\midrule
$w_A$ (motion weight) & 0.3 & 0.794 & 0.760 & 0.0613 \\
& 0.4 & 0.803 & 0.770 & 0.0568 \\
& \textbf{0.5} & \textbf{0.802} & \textbf{0.776} & \textbf{0.0563} \\
& 0.6 & 0.799 & 0.768 & 0.0569 \\
& 0.7 & 0.795 & 0.765 & 0.0571 \\
\midrule
$w_P$ (sphere weight) & 0.1 & 0.798 & 0.765 & 0.0589 \\
& 0.2 & 0.803 & 0.778 & 0.0567 \\
& \textbf{0.3} & \textbf{0.802} & \textbf{0.776} & \textbf{0.0563} \\
& 0.4 & 0.803 & 0.759 & 0.0570 \\
& 0.5 & 0.801 & 0.748 & 0.0583 \\
\midrule
$w_D$ slope & 0.0 & 0.803 & 0.776 & 0.0569 \\
& 0.1 & 0.807 & 0.775 & 0.0553 \\
& \textbf{0.15} & \textbf{0.802} & \textbf{0.776} & \textbf{0.0563} \\
& 0.2 & 0.801 & 0.771 & 0.0564 \\
\midrule
$\tau_c$ (overlap onset) & 0.2 & 0.803 & 0.775 & 0.0559 \\
& \textbf{0.3} & \textbf{0.802} & \textbf{0.776} & \textbf{0.0563} \\
& 0.4 & 0.802 & 0.768 & 0.0565 \\
\midrule
$\alpha/\beta$ (SNR exp.) & 0.4/0.1 & 0.802 & 0.764 & 0.0578 \\
& \textbf{0.6/0.2} & \textbf{0.802} & \textbf{0.776} & \textbf{0.0563} \\
& 1.0/0.5 & 0.799 & 0.773 & 0.0562 \\
& 2.0/1.0 & 0.789 & 0.757 & 0.0585 \\
\bottomrule
\end{tabular}
\caption{Reliability-score sensitivity analysis.
Each block varies one hyperparameter while holding all others at their
default (\textbf{bold}). Results on the IA-Bench test split. All values are
fixed across datasets and embodiments. This is a post-hoc diagnostic: it was run after the defaults were frozen and was not used to select them.}
\label{tab:sensitivity}
\end{table}

\subsection{Weight Selection and Learned Combiner}
\label{app:weights}
The default weights were set manually on the validation split, guided by the intended role of each cue, and frozen before any test evaluation.
To test whether results depend on this choice, we evaluate the reliability score on the 443 emitted validation samples (5,695 candidates) under three sweeps: varying one parameter at a time, all 243 joint combinations of neighboring values, and all 1,200 joint combinations of the parameter values in Table~\ref{tab:sensitivity} (Table~\ref{tab:weight_selection}a).
72.4\% of the local configurations remain within one accuracy point of the default.
We then select a single configuration by minimum validation AURC ($w_A{=}.4$, $w_P{=}.3$, $w_D$ slope $.2$, $\tau_c{=}.2$, $\alpha/\beta{=}.6/.2$) and evaluate it once on the test split.
As a learned alternative, we fit a plain logistic regression on standardized candidate-level features ($A$, $D$, $P$, $1-\Pi$) to predict whether a candidate matches the ground truth, using trajectory-grouped cross-validation on the same validation samples, without class or sample weighting or engineered interactions.
OXE is never used for fitting, model selection, or validation.
Table~\ref{tab:weight_selection}b shows that joint tuning changes test results only marginally, while the learned combiner is worse in accuracy, AURC, and Cov@90, with the largest gap on zero-shot OXE.
Unlike the fixed rule, a linear combiner cannot express the detector weight that adapts to proximity or the depth-relieved overlap penalty.

\begin{table}[H]
\centering
\scriptsize
\setlength{\tabcolsep}{3.1pt}
\renewcommand{\arraystretch}{1.06}
\textbf{(a) Validation sensitivity}\\[2pt]
\begin{tabular}{lrrrr}
\toprule
Sweep & Configs & Acc. range & AURC range & Cov@90 range \\
\midrule
Fixed default & 1 & .847 & .0321 & .867 \\
One-at-a-time & 17 & .819--.851 & .0312--.0427 & .795--.876 \\
Joint local & 243 & .815--.853 & .0305--.0432 & .804--.892 \\
Joint broad & 1,200 & .783--.856 & .0305--.0589 & .716--.892 \\
\bottomrule
\end{tabular}

\vspace{6pt}
\textbf{(b) Held-out test comparison}\\[2pt]
\begin{tabular}{lrrrrrr}
\toprule
& \multicolumn{3}{c}{All test ($n=1{,}256$)}
& \multicolumn{3}{c}{OXE test ($n=424$)} \\
\cmidrule(lr){2-4}\cmidrule(lr){5-7}
Method & Acc. & AURC & Cov@90 & Acc. & AURC & Cov@90 \\
\midrule
Logistic regression & .789 & .0607 & .750 & .809 & .0449 & .849 \\
Validation-selected weights & \textbf{.803} & \textbf{.0560} & \textbf{.783} & .830 & \textbf{.0376} & \textbf{.875} \\
Fixed \gls{method} & .802 & .0563 & .776 & \textbf{.835} & .0387 & .863 \\
\bottomrule
\end{tabular}
\caption{\textbf{Weight selection and learned combination.} (a) Sweeps on the validation split. (b) Test results of the fixed default, the configuration selected by minimum validation AURC, and a logistic-regression combiner trained on the validation split.}
\label{tab:weight_selection}
\end{table}

\subsection{Cue Analysis}
\label{app:cue_analysis}
Table~\ref{tab:margin_ladder} reports the incremental accuracy and Cov@90 gain of each stage of the ablation in Table~\ref{tab:main_results}, stratified by the detector's confidence margin between its top two candidates.
Motion contributes most where appearance is ambiguous, and proximity mainly improves ranking in the low-margin tercile, adding 13.4 Cov@90 points but only 0.2 accuracy points.
Motion and proximity are partly redundant (Spearman $\rho=.406$), which explains the small aggregate effect of removing proximity.
Separating the two algebraic roles of proximity shows that the exact adaptive detector weight is not critical: direct proximity with a fixed detector weight obtains .803 accuracy and .0569 AURC, the full adaptive rule .802/.0563, the adaptive weight without the direct term .791/.0640, and neither .795/.0612.

The conditional cues matter in their intended regimes.
On the 72 test samples whose ground-truth object overlaps the robot mask by at least 0.3, removing depth relief lowers accuracy from .694 to .514 and Cov@90 from .472 to .056 (paired $p=.001$), and removing proximity lowers accuracy to .611 ($p=.070$).
On the 116 single-phase fallback samples, removing proximity lowers accuracy from .741 to .707 ($p=.344$).

\begin{table}[H]
\centering
\small
\setlength{\tabcolsep}{5pt}
\begin{tabular}{lrrr}
\toprule
Added stage & Low margin & Medium margin & High margin \\
\midrule
Mean motion        & $+24.1/+14.8$ & $+10.8/+24.4$ & $+0.0/+0.0$ \\
Robot filtering    & $+6.4/+0.0$   & $+3.1/+9.1$   & $-0.7/+0.0$ \\
Soft-SNR shaping   & $+5.7/+8.4$   & $+7.4/+23.0$  & $+2.1/+0.5$ \\
3D proximity       & $+0.2/+13.4$  & $+1.9/+6.7$   & $+0.0/-0.7$ \\
Soft overlap/depth & $+1.2/+0.5$   & $+1.4/+1.4$   & $+2.4/+1.9$ \\
\bottomrule
\end{tabular}
\caption{\textbf{Gains by appearance ambiguity.} Incremental accuracy/Cov@90 change in percentage points for each ablation stage, stratified into terciles of the detector's top-two confidence margin. The full score raises low-margin accuracy from .358 to .735.}
\label{tab:margin_ladder}
\end{table}

Table~\ref{tab:task_family_cues} shows which cues help which task families.
Phase-aware motion is the most important cue for relocation, open/close, and fluid tasks, and robot overlap matters most for non-prehensile tasks, where the gripper and the object move together.
The 13 button and press samples reveal a limitation: the contact is visually static, so motion and depth relief become unreliable and removing them even improves accuracy.
For such tasks, contact events from force-torque or tactile sensing, or end-effector deceleration, could replace visual grasp phases.

\begin{table}[H]
\centering
\scriptsize
\setlength{\tabcolsep}{3pt}
\begin{tabular}{lrrrrrrr}
\toprule
Task family ($n$) & Det. & Full & $\Delta{-}A$ & $\Delta{-}D$ & $\Delta{-}P$ & $\Delta{-}$overlap & $\Delta{-}$relief \\
\midrule
Relocation (884)        & .629 & .861 & $-.189$ & $-.045$ & $-.008$ & $-.054$ & $-.014$ \\
All non-prehensile (372) & .468 & .661 & $-.121$ & $-.035$ & $-.003$ & $-.099$ & $+.024$ \\
Open/close (117)        & .368 & .581 & $-.103$ & $+.000$ & $-.009$ & $-.111$ & $+.017$ \\
Fluid/wipe (49)         & .388 & .857 & $-.429$ & $-.041$ & $-.020$ & $-.184$ & $+.000$ \\
Tool/shape (48)         & .854 & .896 & $+.000$ & $+.000$ & $+.021$ & $-.042$ & $+.000$ \\
Button/press (13)       & .308 & .385 & $+.308$ & $-.231$ & $+.000$ & $-.154$ & $+.308$ \\
\bottomrule
\end{tabular}
\caption{\textbf{Cue importance by task family.} $\Delta{-}X$ is the accuracy change after removing cue $X$ from the full score. Negative values mean the cue helps.}
\label{tab:task_family_cues}
\end{table}

\subsection{EgoLive Transfer}
\label{app:egolive}
EgoLive~\cite{li2026egolive} contains human demonstrations recorded with a head-mounted device during real tasks in homes, retail, and care settings.
We use the LeRobot v3 release and the left head camera, downsampled from 2160$\times$2160 at 60\,fps to 512$\times$512 at 10\,fps.
Each trajectory is one of EgoLive's official instruction segments, widened by 0.5\,s on either side so that the clip shows the hand before and after contact. In total, we evaluate 733 segments.
Hand boxes, hand-object boxes, 2D/3D hand keypoints, and per-frame camera poses are provided by the dataset and replace the proprioceptive signals available for robots.
Interaction phases are derived from the per-frame hand-object boxes, which we temporally smooth because the raw contact signal flickers. Since the dataset's track identities do not persist, objects are matched across frames by box overlap.

\textbf{Hand masks and proximity.}
Hand masks replace the robot segmentation. We run SAM2 on 8 keyframes per phase, prompted with the hand box, that hand's 21 keypoints as positive points, and the other hand's keypoints as negative points, so that touching hands are not merged.
A mask is required because a held object lies inside the hand box, so a box-based overlap would count the correct object as part of the hand.
Hand proximity replaces the 3D gripper sphere: the grasp center is the midpoint of the thumb and index fingertips, and proximity is the fraction of a candidate's visible tracked points within twice the half-diagonal of the hand box from this point.
The radius thus scales with the hand's image size and requires no metric depth. Since we do not run depth estimation, the depth relief of the overlap penalty is dropped, which makes the penalty stricter.

\textbf{Object movement under camera motion.}
A candidate's position in each frame is the mean of its visible tracked points. We first remove head rotation exactly by mapping each frame to the first frame's camera orientation with the homography $H=KR_0^\top R_kK^{-1}$. The remaining shared motion is mostly parallax from head translation, which we remove by subtracting the per-frame median displacement across candidates, assuming that most candidates are static. This step is skipped for fewer than three candidates.
Movement is measured in pixels per second, and a frame counts as interaction if either hand touches the object, so that handovers are not treated as background motion.

\begin{table}[H]
\centering
\small
\begin{tabular}{lccc}
\toprule
Method & Acc.$\uparrow$ & AURC$\downarrow$ & Cov@90$\uparrow$ \\
\midrule
Detector confidence & .492 & .317 & .150 \\
\quad + object movement & .541 & .249 & .274 \\
\quad + hand proximity (\gls{method}) & \textbf{.754} & \textbf{.092} & \textbf{.575} \\
\bottomrule
\end{tabular}
\caption{\textbf{EgoLive transfer} on 733 egocentric human demonstrations. Cues are added cumulatively to detector confidence.}
\label{tab:egolive}
\end{table}

\subsection{IoU-threshold Robustness}
\label{app:iou_thresh}
\begin{table}[t]
\centering
\small
\begin{tabular}{lccc}
\toprule
Method & IoU\,$>$\,0.4 & IoU\,$>$\,0.5 & IoU\,$>$\,0.75 \\
\midrule
Det.~conf.~only & 0.581 & 0.572 & 0.564 \\
+Mean motion & 0.697 & 0.689 & 0.681 \\
+Robot filter (hard) & 0.727 & 0.716 & 0.706 \\
+Soft-SNR motion & 0.778 & 0.766 & 0.759 \\
+Adaptive det.~(sphere) & 0.785 & 0.776 & 0.768 \\
\midrule
SPARC (ours) & \textbf{0.802} & \textbf{0.793} & \textbf{0.783} \\
\bottomrule
\end{tabular}
\caption{Accuracy at stricter IoU match thresholds.
Correctness at each threshold: IoU\,$>$\,threshold, \emph{or} predicted
box $\geq$80\% contained within GT box with IoU\,$>$\,0.1 (containment
fallback identical across columns). SPARC retains its margin at all thresholds.}
\label{tab:iou_robustness}
\end{table}

\subsection{Robot-overlap Filter False-Suppression Analysis}
\label{app:robot_overlap}
\begin{table}[t]
\centering
\small
\begin{tabular}{lrr}
\toprule
Statistic & Count & \% of solvable \\
\midrule
Solvable demonstrations & 1171 & 100.0 \\
GT candidate hard-suppressed ($\geq 0.3$ overlap) & 72 & 6.1 \\
\quad , of which false suppressions & 34 & 2.9 \\
True suppressions (wrong\,$\to$\,correct) & 60 & 5.1 \\
\midrule
Acc without filter (+Mean motion) & \multicolumn{2}{c}{0.697} \\
Acc with hard filter & \multicolumn{2}{c}{0.727} \\
Acc SPARC (quadratic penalty, ours) & \multicolumn{2}{c}{\textbf{0.802}} \\
\bottomrule
\end{tabular}
\caption{Robot-overlap filter false-suppression analysis on the IA-Bench
test split.
Correctness uses the IoU+containment criterion throughout.
``GT hard-suppressed'' counts demonstrations where the correct candidate
received overlap $\geq 0.3$ and was gated to $-\infty$ in the ablation
row \emph{+Robot filter (hard)}.
``False suppression'' is the subset where \emph{+Mean motion} (no filter)
would have selected the correct candidate.
SPARC's quadratic penalty avoids hard gating and lifts accuracy from
0.727 to 0.802.}
\label{tab:filter_analysis}
\end{table}

\subsection{Per Dataset Results}
\label{sec:per_dataset_results}
Table~\ref{tab:per_dataset} reports the full ablation across all four IA-Bench datasets.
FSD abstentions are counted as incorrect throughout. Agibot proves to be the most challenging dataset, compared to Bridge, which consists mostly of simple pick-and-place in visually simple environments.
\begin{table}[H]
  \centering
  \small
  \setlength{\tabcolsep}{3.5pt}
  \renewcommand{\arraystretch}{1.12}
  \begin{tabular}{l l cccccc}   
  \toprule
  \textbf{Dataset} & \textbf{Scoring rule} &
  \textbf{Acc.}$\uparrow$ &
  \textbf{Cov@90}$\uparrow$ &
  \textbf{Cov@95}$\uparrow$ &
  \textbf{AURC}$\downarrow$ &
  \textbf{E-AURC}$\downarrow$ \\
  \midrule  
  \multirow{6}{*}{\textit{Bridge} ($N$=311)}
  & FSD~\cite{yuan2025fsd}                  & 0.598 & 0.624 & 0.460 & 0.113 & 0.018 \\
  & Det.\ confidence                        & 0.698 & 0.486 & 0.347 & 0.123 & 0.071 \\
  & \quad + Mean movement                   & 0.826 & 0.711 & 0.450 & 0.060 & 0.044 \\
  & \quad + Robot filter                    & 0.871 & 0.894 & 0.643 & 0.041 & 0.032 \\
  & \quad + Phase aware movement            & 0.907 & 1.000 & 0.894 & 0.021 & 0.016 \\
  & \cellcolor{black!5}\textbf{Ours (Final)}& \cellcolor{black!5}\textbf{0.913} & \cellcolor{black!5}\textbf{1.000} & \cellcolor{black!5}\textbf{0.904} & \cellcolor{black!5}\textbf{0.020} & \cellcolor{black!5}\textbf{0.016} \\
  \midrule
  \multirow{6}{*}{\textit{AgiBotWorld} ($N$=258)}
  & FSD~\cite{yuan2025fsd}                  & 0.357 & 0.225 & 0.116 & 0.330 & 0.053 \\
  & Det.\ confidence                        & 0.450 & 0.039 & 0.008 & 0.376 & 0.184 \\
  & \quad + Mean movement                   & 0.597 & 0.271 & 0.167 & 0.222 & 0.127 \\
  & \quad + Robot filter                    & 0.578 & 0.275 & 0.159 & 0.223 & 0.117 \\
  & \quad + Phase aware movement            & 0.655 & 0.360 & 0.236 & 0.153 & 0.084 \\
  & \cellcolor{black!5}\textbf{Ours (Final)}& \cellcolor{black!5}\textbf{0.674} & \cellcolor{black!5}\textbf{0.372} & \cellcolor{black!5}\textbf{0.244} & \cellcolor{black!5}\textbf{0.145} & \cellcolor{black!5}\textbf{0.085} \\
  \midrule
  \multirow{6}{*}{\textit{DROID} ($N$=263)}
  & FSD~\cite{yuan2025fsd}                  & 0.456 & 0.437 & 0.335 & 0.225 & 0.038 \\
  & Det.\ confidence                        & 0.563 & 0.000 & 0.000 & 0.267 & 0.152 \\
  & \quad + Mean movement                   & 0.646 & 0.498 & 0.338 & 0.131 & 0.059 \\
  & \quad + Robot filter                    & 0.692 & 0.494 & 0.338 & 0.121 & 0.067 \\
  & \quad + Phase aware movement            & 0.741 & 0.654 & 0.570 & 0.077 & 0.039 \\
  & \cellcolor{black!5}\textbf{Ours (Final)}& \cellcolor{black!5}\textbf{0.741} & \cellcolor{black!5}\textbf{0.654} & \cellcolor{black!5}\textbf{0.570} & \cellcolor{black!5}\textbf{0.077} & \cellcolor{black!5}\textbf{0.040} \\
  \midrule
  \multirow{6}{*}{\textit{OXE} ($N$=424)}
  & FSD~\cite{yuan2025fsd}                  & 0.229 & 0.250 & 0.208 & 0.439 & 0.004 \\
  & Det.\ confidence                        & 0.587 & 0.380 & 0.302 & 0.192 & 0.091 \\
  & \quad + Mean movement                   & 0.696 & 0.557 & 0.486 & 0.110 & 0.058 \\
  & \quad + Robot filter                    & 0.733 & 0.623 & 0.524 & 0.083 & 0.044 \\
  & \quad + Phase aware movement            & 0.802 & 0.828 & 0.757 & 0.043 & 0.022 \\
  & \cellcolor{black!5}\textbf{Ours (Final)}& \cellcolor{black!5}\textbf{0.835} & \cellcolor{black!5}\textbf{0.863} & \cellcolor{black!5}\textbf{0.762} & \cellcolor{black!5}\textbf{0.039} & \cellcolor{black!5}\textbf{0.024} \\
  \bottomrule
  \end{tabular}
  \caption{\textbf{Per-dataset ablation on IA-Bench.}
  Each ablation stage is applied cumulatively from detection confidence alone up to the full SPARC score.
  FSD abstentions (53\% of samples on OXE) are counted as incorrect.
  Cov@90/95 = coverage at 90\%/95\% precision operating point.}
  \label{tab:per_dataset}
  \end{table}

\subsection{Per Task Results}

\begin{table}[t]
  \centering
  \caption{Accuracy broken down by task group across all datasets (IA-Bench
    test split, $n{=}1{,}256$). Correctness uses the IoU+containment
    criterion. FSD abstains on 53\% of demonstrations. Abstentions are
    counted as incorrect, consistent with the requirement of full-corpus
    annotation coverage.}
  \label{tab:per_task}
  \small
  \setlength{\tabcolsep}{4pt}
  \begin{tabular}{lc ccccc c}
  \toprule  
   & & \multicolumn{6}{c}{Method} \\
  \cmidrule(lr){3-8}
  Task group & $n$ & Det.\ conf. & +Motion & +Filter & +Soft-SNR & SPARC & FSD \\
  \midrule
  Relocate         & 884 & 0.629 & 0.761 & 0.776 & 0.828 & \textbf{0.861} & 0.436 \\
  Open\,/\,Close   & 117 & 0.368 & 0.530 & 0.590 & 0.615 & \textbf{0.581} & 0.222 \\
  Articulate       &  47 & 0.574 & 0.532 & 0.596 & \textbf{0.681} & 0.638 & 0.340 \\
  Press\,/\,Switch &  13 & 0.308 & 0.154 & 0.231 & 0.385 & \textbf{0.385} & 0.000 \\
  Fluid            &  49 & 0.388 & 0.592 & 0.735 & 0.776 & \textbf{0.857} & 0.306 \\
  Tool\,/\,Shape   &  48 & 0.854 & 0.896 & 0.896 & \textbf{0.917} & 0.896 & 0.625 \\
  Other            &  98 & 0.408 & 0.429 & 0.490 & 0.551 & \textbf{0.592} & 0.235 \\
  \midrule
  \textbf{All}     & \textbf{1256} & 0.581 & 0.697 & 0.727 & 0.778 & \textbf{0.802} & 0.394 \\
  \bottomrule
  \end{tabular}
  \end{table}
\label{app:per_task_results}
We further show the performance of \gls{method} on different task groups.

Task groups are derived from the action label produced by the LLM subtask
parser (Sec.~\ref{sec:method}): we match the action string against keyword
lists for each group (e.g.\ \emph{pick}, \emph{place}, \emph{lift}
$\to$ Relocate, and \emph{open}, \emph{close} $\to$ Open/Close) and assign
unmatched actions to \emph{Other}.

  SPARC achieves its strongest results on the two categories that have the clearest motion
  signal: \emph{relocate} (0.861) and \emph{fluid} (0.857).
  Both involve sustained, large-amplitude end-effector displacement that the SNR-based motion
  score is well-suited to capture.
  \emph{Tool/shape} tasks (0.896) benefit additionally from the dedicated tool-detection branch,
  which explicitly segments the held instrument and transfers its tracks to the manipulated object.
  Performance is weakest on \emph{press/switch} (0.385) and \emph{open/close} (0.581),
  tasks characterized by small spatial displacement or purely rotational motion. Here, the
  motion signal carries less discriminative power, and the sphere-shape prior is rarely applicable.
  \emph{Articulate} tasks show a modest regression when the sphere bonus is added,
  suggesting that the spherical-object heuristic occasionally fires on non-articualted
  articulated objects such as drawer handles.
  Across every group, SPARC substantially outperforms FSD\@.
  FSD's coverage-adjusted accuracy collapses to 0.000 on press/switch because nearly all
  of those objects are small and briefly tracked, causing the area-and-duration filter to
  abstain on the entire group.
  This confirms that a purely detector-driven, coverage-agnostic baseline cannot generalise
  across the diversity of manipulation task types present in IA-Bench.

\section{Downstream Embodied VLM Training}
\subsection{Hyperparameters}
\begin{table*}[h]
\centering
\tiny
\setlength{\tabcolsep}{5pt}
\begin{tabular}{lcccc}
\toprule
Hyperparameter &
Ours + Mix + Threshold &
Det + Mix + Threshold &
Mix (FSD+RoboPoint+OV2) &
EO-1.5M + Mix \\

\midrule
Model & Qwen3.5-4B & Qwen3.5-4B & Qwen3.5-4B & Qwen3.5-4B \\
Per-device batch size & 8 & 8 & 8 & 8 \\
Gradient accumulation & 1 & 1 & 1 & 1 \\
Effective global batch size & 128 & 128 & 128 & 128 \\
Epochs & 1 & 1 & 1 & 1 \\
Learning rate & $2\times10^{-5}$ & $2\times10^{-5}$ & $2\times10^{-5}$ & $2\times10^{-5}$ \\
LR scheduler & cosine w/ min LR & cosine w/ min LR & cosine w/ min LR & cosine w/ min LR \\
Warmup ratio & 0.03 & 0.03 & 0.03 & 0.03 \\
Min LR ratio & 0.1 & 0.1 & 0.1 & 0.1 \\
Weight decay & 0.0 & 0.0 & 0.0 & 0.0 \\
Max grad norm & 1.0 & 1.0 & 1.0 & 1.0 \\
Precision & bf16 & bf16 & bf16 & bf16 \\
Gradient checkpointing & true & true & true & true \\
Vision encoder frozen & true & true & true & true \\
Vision projector frozen & false & false & false & false \\
Max sequence length & 5600 & 5600 & 5600 & 5600 \\
\midrule
Candidate selector &
\gls{method} &
Detection Confidence &
-- &
-- \\
Max samples &
all &
all &
all &
all \\
Quality threshold & 0.97 & 0.85 & -- & -- \\
Max per object & 700 & 700 & -- & -- \\
Data mix &
\makecell{Ours + fsd +\\ robopoint + llavaov2} &
\makecell{Det + fsd +\\ robopoint + llavaov2} &
\makecell{fsd + robopoint +\\ llavaov2} &
\makecell{EO-1.5M + fsd +\\ robopoint + llavaov2} \\
Dataset Size &
1,159,047 &
941,642 &
793,580  &%
1,063,630 \\
\bottomrule
\end{tabular}
\caption{Training hyperparameters for the main mixture-based models used in our downstream VLM training experiments.}
\label{tab:appendix_training_hparams}
\end{table*}
\subsection{VQA Dataset Generation}
\label{app:vqa_dataset}

To generate the VQA dataset, we use templated question-answer pairs. We consider three task types. In object grounding, the VLM points to a specified object from either its name or the demonstration instruction. In vacant location pointing, the VLM points to the start or target location of the demonstration. In trajectory prediction, the VLM predicts the object trajectory for a given task instruction. We use the same templates and prompts for all variants, and only replace the underlying spatial annotations with either \gls{method} or detection-generated annotations.

For target point supervision, we sample a point from the winner candidate mask. For vacant pointing, we use the extracted start or target location and query the first or last observation frame in which the object is absent at that location. We then sample points at the corresponding vacant location. This provides a proxy for pointing to a vacant location without requiring explicit location annotations, thereby allowing supervision from the language instruction alone. For trajectory tasks, we subsample five equally spaced points from the object trajectory. All spatial coordinates are normalized to the range $[0,1000]$.

\subsection{Downstream Dataset Scaling Experiments}
\label{app:scaling}

We study how annotation selection quality affects downstream scaling (Figure~\ref{fig:sparc_scaling_combined}). For each scoring rule, we rank automatically generated annotations and train models on the top-scoring 50K, 200K, 500K, and 838,211 samples. We do not mix other datasets with the respective data subsets. \gls{method} ranks samples using the proposed interaction-aware reliability score, while the baseline ranks samples by detector confidence. We additionally report a quality-filtered reference for each method using the same fixed quality threshold as in the main experiments.

SPARC provides a stronger scaling signal than detector confidence on both the overall benchmark suite and IA-Bench. On the aggregate score, SPARC improves from 0.554 at 50K to 0.583 at 200K, then decreases to 0.563 at 500K and 0.557 at 838K. This suggests that scaling is beneficial until lower-ranked annotations enter a noisier data regime. Note that the drop in performance on other general benchmarks can also be attributed to the model overfitting on our domain when not mixing in general VQA data. However,  on \gls{bench}, SPARC continues to improve with scale, rising from 0.716 at 50K to 0.785 at 838K, close to the quality-filtered reference of 0.789. In contrast, detector confidence degrades as more samples are added in both settings. This indicates that detector confidence increasingly admits noisy supervision, while SPARC preserves useful supervision over a much larger annotation budget.

\subsection{Comparison to Embodied Foundation Models}
\begin{table*}[h]
\centering
\scriptsize
\setlength{\tabcolsep}{3pt}
\renewcommand{\arraystretch}{1.2}
\resizebox{\textwidth}{!}{%
\begin{tabular}{
l
*{9}{c}
@{\hspace{1.5em}} 
*{5}{c}
}
\toprule
&
\multicolumn{9}{c}{\textbf{Pointing and VQA} $\uparrow$}
&
\multicolumn{5}{c}{\textbf{Trajectory benchmarks} $\downarrow$} \\
\cmidrule(lr){2-10}
\cmidrule(l){11-15}
\textbf{Method}
& \makecell{IA \\Bench}
& \makecell{Where2\\Place}
& \makecell{Ref\\Spatial}
& ERQA
& \makecell{Robo\\Spatial}
& \makecell{Robo\\RefIt}
& \makecell{EO\\Bench}
& \makecell{VA\\Bench-P}
& \makecell{\textbf{Avg.}}
& \makecell{RoboInter\\Gripper}
& \makecell{RoboInter\\Traj.}
& \makecell{ShareRobot\\Bench-T}
& \makecell{VA\\Bench-V}
& \makecell{\textbf{Avg.}} \\
\midrule

Qwen3.5-4B (Zero-shot)
& 49.7 & 47.0 & 30.5 & 30.3 & 48.4 & 70.1 & 30.5 & 19.0 & 42.6
& 0.180 & 0.282 & $\infty$ & 0.219 & $\infty$ \\

\midrule
\multicolumn{15}{l}{\textit{Open Source Brain Models}} \\
\midrule
RoboBrain 2.0 (7B)
  & -         %
  & 63.5         %
  & 54.0  %
  & 30.3   %
  & 54.2   %
  & 70.4         %
  & -         %
  & 26.67         %
  & -         %
  & -         %
  & -         %
  & 0.236         %
  & -         %
  & -         %
  \\

MiMo-Embodied (7B)
  & -         %
  & 63.6         %
  & 48.0  %
  & 46.7  %
  & 61.8  %
  & 82.3         %
  & -         %
  & 46.9         %
  & -         %
  & -         %
  & -         %
  & -         %
  & -         %
  & -         %
  \\

RoboPoint (13B)
  & -         %
  & 46.8  %
  & 16.1  %
  & -         %
  & -         %
  & 49.8  %
  & -         %
  & 19.1  %
  & -         %
  & -         %
  & -         %
  & -         %
  & -         %
  & -         %
  \\

RoboRefer (2B SFT)
  & -         %
  & 66.0  %
  & 33.8  %
  & -         %
  & \textbf{66.4}  %
  & 72.8  %
  & -         %
  & 24.7  %
  & -         %
  & -         %
  & -         %
  & -         %
  & -         %
  & -         %
  \\

Gemini 2.5 Pro
  & -         %
  & 53.0  %
  & 29.2  %
  & 48.3  %
  & -         %
  & 49.5  %
  & -         %
  & 21.7  %
  & -         %
  & -         %
  & -         %
  & -         %
  & -         %
  & -         %
  \\

Qwen2.5-VL (72B)
  & -         %
  & 37.2 %
  & 20.9  %
  & -         %
  & -         %
  & 78.5  %
  & -         %
  & 23.3  %
  & -         %
  & -         %
  & -         %
  & -         %
  & -         %
  & -         %
  \\

EO-1 (3B)
  & -         %
  & -         %
  & -         %
  & 45.5  %
  & -         %
  & -         %
  & 36.4  %
  & -         %
  & -         %
  & -         %
  & -         %
  & -         %
  & -         %
  & -         %
  \\

FSD (13B)~\cite{yuan2025fsd}
  & -         %
  & 45.8      %
  & -         %
  & -         %
  & -         %
  & 56.7      %
  &   %
  & 61.8  %
  & -         %
  & -         %
  & -         %
  & -         %
  & -        %
  & -         %
  \\
RynnBrain (8B)
  & -         %
  & -         %
  & 59.2      %
  & 46.8      %
  & \textbf{73.1}      %
  & -         %
  & -         %
  & -         %
  & -      %
  & -         %
  & -         %
  & 0.35      %
  & -         %
  & -         %
  \\
RynnBrain (2B)
  & -         %
  & -         %
  & 52.7      %
  & 42.3      %
  & 65.7      %
  & -         %
  & -         %
  & -         %
  & -     %
  & -         %
  & -         %
  & 0.34      %
  & -         %
  & -         %
  \\
\midrule
\rowcolor{gray!15}
\textbf{Ours} (4B)
& \textbf{79.1} & \textbf{71.0} & \textbf{54.5} & \textbf{44.9} & \textbf{61.0} & \textbf{86.4} & 35.3 & \textbf{65.7} & \textbf{62.7}
& \textbf{0.125} & 0.319 & \textbf{0.232} & \textbf{0.091} & \textbf{0.192} \\
\rowcolor{gray!15}
\textbf{Ours} (0.8B, VT-FT)
& 76.7 & 58.0 & 37.1 & 40.4 & 53.5 & 81.0 & 30.3 & 48.3 & 53.2
& 0.175 & 0.429 & 0.354 & 0.216 & 0.294 \\

\rowcolor{gray!15}
\textbf{Ours} (0.8B)
& 75.4 & 62.0 & 33.1 & 36.9 & 45.5 & 81.0 & 34.6 & 56.3 & 53.1
& 0.157 & 0.425 & 0.358 & 0.208 & 0.287 \\

\bottomrule

\end{tabular}%
}
\caption{
Comparison of Qwen3.5-4B trained on \gls{method} data against several recent state-of-the-art embodied foundation models. All values are taken from the respective papers. Although our VLM is smaller in size and trained on vastly less data without any human annotation, it remains competitive with larger models while even outperforming most of them consistently across several benchmarks.
}
\label{tab:downstream_brain_models}
\end{table*}

\clearpage
\section{Real-Robot Setup and Policy Training}

\subsection{Real Robot Setup and Policy Training}

\begin{figure}[h]
    \centering
    \includegraphics[
        width=0.4\linewidth,
        trim=0 800 0 1000,
        clip
    ]{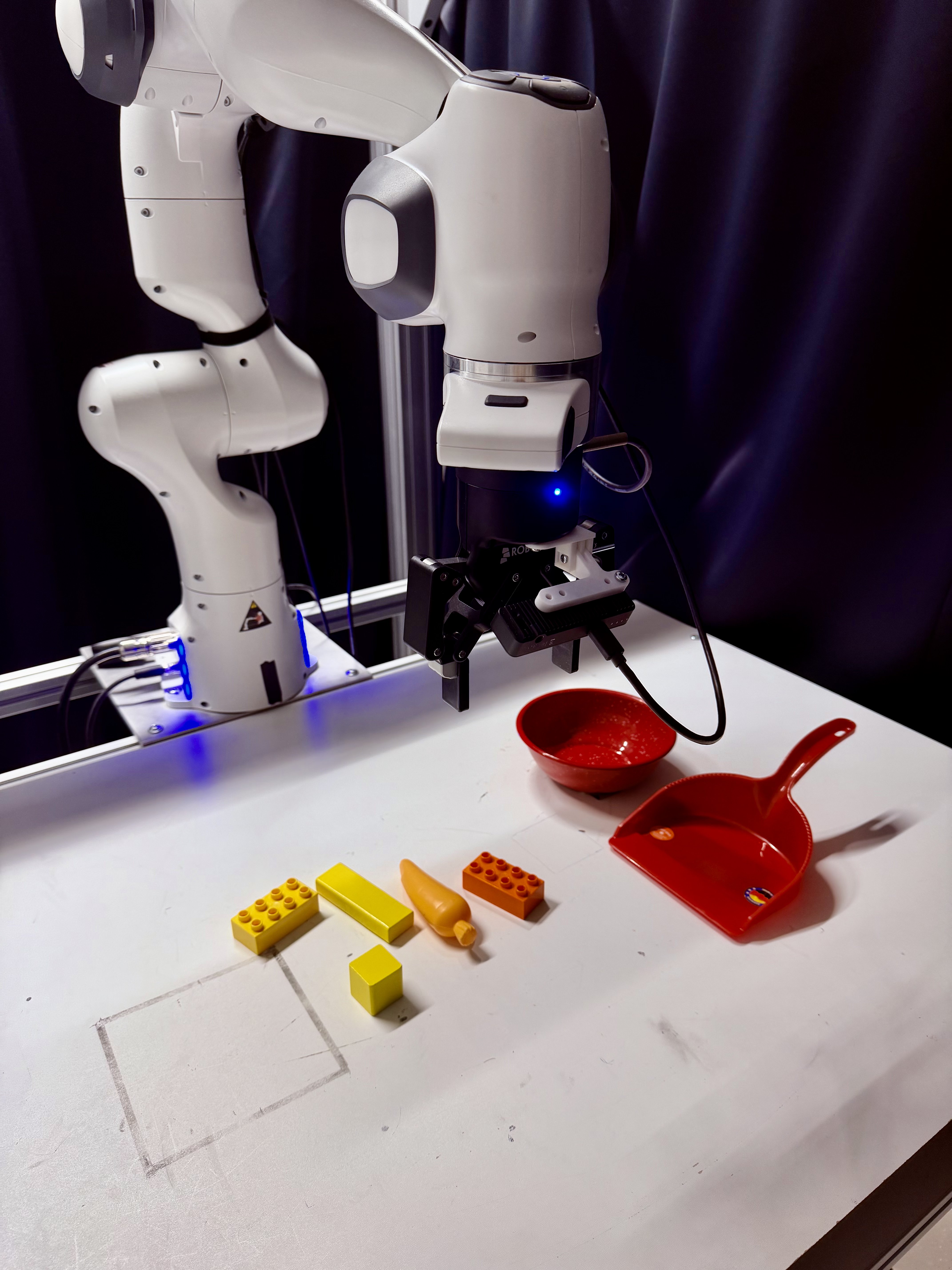}
    \caption{Illustration of our real-world robot evaluation setup.}
    \label{fig:real_robot}
\end{figure}

We conduct our real robot experiments in a tabletop manipulation setting. Specifically, we use a Franka-Panda manipulator with a Robotiq gripper and one external and one in-hand camera. We conduct the 10 tasks shown in \ref{tab:ambiguity_instructions}. The setup and the objects the robot has to manipulate are shown in \Cref{fig:real_robot}.

For policy training, we employ the following logic:
For each trajectory $(o_i, \ell_i, a_i)$, we optionally retrieve a step-aligned reasoning trace $c_i$ either generated from the detection baseline or \gls{method} and train the policy to produce this reasoning trace before predicting the low-level action.  The reasoning trace is generated and filtered with the respective method (\gls{method} and Detection Baseline). We simplify each object's trajectory by uniformly resampling it into 5 waypoints along its arc length.
This preserves the coarse geometric path while removing redundant high-frequency tracking samples.  The trace is prompted as:
\emph{``Generate the 2D trajectory the object should follow to complete the task. Output exactly 5 points.''}
The VLM is supervised only on the assistant reasoning tokens using a masked language-modeling loss,
\begin{equation}
\mathcal{L}_{\mathrm{cot}}
=
-\frac{1}{|\mathcal{M}|}
\sum_{i}\sum_{t\in\mathcal{M}_i}
\log p_{\theta}(c_{i,t}\mid o_i,\ell_i,c_{i,<t}),
\end{equation}
where $\mathcal{M}_i$ denotes the unmasked CoT answer tokens. The action expert is a diffusion-transformer policy that predicts a horizon of $H=24$ continuous 8-DoF absolute end-effector actions,
\begin{equation}
a_i = (a_i^1,\ldots,a_i^H), \qquad a_i^t \in \mathbb{R}^{8}.
\end{equation}
It conditions on the last VLM hidden states $h_i$ and learns to denoise noisy action trajectories. 
The final objective is
\begin{equation}
\mathcal{L}
=
\mathcal{L}_{\mathrm{act}}
+
0.1\,\mathcal{L}_{\mathrm{cot}}.
\end{equation}
At inference time, we first generate the reasoning trace using at most $160$ new tokens, then re-encode the prompt and generated trace to obtain $h_i$ and condition the action expert.\\
Cotraining with good spatial annotations gives the VLMs the capability to ground visually very similar objects. Notably, training on low quality annotations results in a significant performance drop.
\begin{table}[t]
\centering
\small
\setlength{\tabcolsep}{8pt}
\renewcommand{\arraystretch}{1.1}
\begin{tabular}{ll}
\toprule
\textbf{ID} & \textbf{Language instruction} \\
\midrule
1  & Put the yellow cube into the red bowl \\
2  & Put the yellow cube into the red dustpan \\
3  & Put the yellow cuboid into the red bowl \\
4  & Put the yellow cuboid into the red dustpan \\
5  & Put the yellow lego block into the red bowl \\
6  & Put the yellow lego block into the red dustpan \\
7  & Put the orange carrot into the red bowl \\
8  & Put the orange carrot into the red dustpan \\
9  & Put the orange lego block into the red bowl \\
10 & Put the orange lego block into the red dustpan \\
\bottomrule
\end{tabular}
\caption{Language instructions used in the real-robot ambiguity evaluation.}
\label{tab:ambiguity_instructions}
\end{table}

\subsection{Per Task Results}
\begin{table}[H]
\centering
\small
\setlength{\tabcolsep}{5pt}
\renewcommand{\arraystretch}{1.15}
\begin{tabular}{lcccccccccc|c}
\toprule
\textbf{Method} 
& \textbf{T1} & \textbf{T2} & \textbf{T3} & \textbf{T4} & \textbf{T5}
& \textbf{T6} & \textbf{T7} & \textbf{T8} & \textbf{T9} & \textbf{T10}
& \textbf{Total} \\
\midrule
Baseline       
& 1 & 2 & 0 & 0 & 0 & 1 & 1 & 0 & 3 & 1 & 9/100 \\
Detection 
& 2 & 1 & 2 & 2 & 0 & 1 & 0 & 0 & 1 & 3 & 12/100 \\
\gls{method}           
& 3 & 7 & 4 & 4 & 1 & 1 & 1 & 2 & 3 & 5 & 31/100 \\
\bottomrule
\end{tabular}
\caption{Per-task real-robot success counts. Each entry denotes successful rollouts out of 10. Two-sided Fisher's exact tests on the pooled counts give $p=.0017$ (\gls{method} vs.\ Detection Reasoning) and $p=.00015$ (\gls{method} vs.\ Baseline).}
\label{tab:real_robot_per_task_counts}
\end{table}

\Cref{tab:real_robot_per_task_counts} shows policy success rates for each tasks. Reasoning VLAs trained on annotations generated by \gls{method} consistently outperform the baselines by a large margin on most tasks, highlighting the improved grounding and spatial localization capabilities resulting from higher quality spatial annotations.

\subsection{Hyperparameters}
\begin{table}[H]
\centering
\small
\resizebox{\textwidth}{!}{
\begin{tabular}{lccc}
\toprule
\textbf{Hyperparameter} &
\textbf{Base} &
\textbf{SPARC Trace CoT} &
\textbf{Detection Trace CoT} \\
\midrule
CoT prompt &
-- &
5-point object trajectory &
5-point object trajectory \\
CoT loss weight &
-- &
$0.1$ &
$0.1$ \\
Generate CoT at inference &
-- &
Yes &
Yes \\
Max CoT tokens &
-- &
$160$ &
$160$ \\
VLM backbone &
Qwen3.5-$0.8$B &
Qwen3.5-$0.8$B &
Qwen3.5-$0.8$B \\
Action expert &
DiT-B &
DiT-B &
DiT-B \\
Action dimension &
$8$ &
$8$ &
$8$ \\
Action horizon &
$24$ &
$24$ &
$24$ \\
Diffusion transformer layers &
$12$ &
$12$ &
$12$ \\
Hidden size &
$768$ &
$768$ &
$768$ \\
Repeated diffusion steps &
$8$ &
$8$ &
$8$ \\
Inference denoising steps &
$4$ &
$4$ &
$4$ \\
Training steps &
$20{,}000$ &
$20{,}000$ &
$20{,}000$ \\
Warmup steps &
$1{,}000$ &
$1{,}000$ &
$1{,}000$ \\
VLA batch size &
$32$ per GPU, $128$ total &
$32$ per GPU, $128$ total &
$32$ per GPU, $128$ total \\
VLM batch size &
$4$ per GPU &
$4$ per GPU &
$4$ per GPU \\
Learning rate, base &
$2.5{\times}10^{-5}$ &
$2.5{\times}10^{-5}$ &
$2.5{\times}10^{-5}$ \\
Learning rate, VLM interface &
$1.0{\times}10^{-5}$ &
$1.0{\times}10^{-5}$ &
$1.0{\times}10^{-5}$ \\
Learning rate, action expert &
$1.0{\times}10^{-4}$ &
$1.0{\times}10^{-4}$ &
$1.0{\times}10^{-4}$ \\
Scheduler &
cosine, min. LR $10^{-6}$ &
cosine, min. LR $10^{-6}$ &
cosine, min. LR $10^{-6}$ \\
Optimizer &
AdamW, $\beta=(0.9,0.95)$ &
AdamW, $\beta=(0.9,0.95)$ &
AdamW, $\beta=(0.9,0.95)$ \\
Weight decay &
$10^{-8}$ &
$10^{-8}$ &
$10^{-8}$ \\
Gradient clipping &
$1.0$ &
$1.0$ &
$1.0$ \\
Gradient accumulation &
$1$ &
$1$ &
$1$ \\
\bottomrule
\end{tabular}
}
\caption{Training hyperparameters for the real-robot models. All models use the same policy and optimization settings. The CoT variants use the same trajectory-style reasoning prompt.}
\label{tab:real_robot_hparams}
\end{table}

\section{Prompts}
\begin{figure*}[h]
\centering

\begin{subfigure}{0.98\textwidth}
\caption{Stage 1a: Single-arm task and object extraction from detected gripper phases}
\begin{lstlisting}[style=promptstyle]
[SYSTEM] You are a robot manipulation expert. Given a task instruction and detected gripper phases with frame indices, extract object interactions (subtasks), assign phases to subtasks, and generate phase descriptions. Each phase has start_frame, end_frame, and phase_type in {"grasp","interact","release","grasp_failure"}. Output a JSON list with keys: object, start_location, target_location, action, tool_name_required, tool_usage_description, grasp_phases. Each grasp_phases item has: start_frame, end_frame, phase_type, description. Use present-tense instruction phrasing. Split multi-object tasks by frame order and task semantics. Keep repeated grasp cycles for the same object in one entry.

[EXAMPLE]
Task: Place the bag of chips in the bottom drawer
Detected phases: [{"start_frame":0,"end_frame":25,"phase_type":"grasp"},{"start_frame":26,"end_frame":80,"phase_type":"interact"},{"start_frame":81,"end_frame":100,"phase_type":"release"}]
Output: {"object":"bag of chips","start_location":null,"target_location":"bottom drawer","action":"pick and place","tool_name_required":null,"tool_usage_description":null,"grasp_phases":[{"start_frame":0,"end_frame":25,"phase_type":"grasp","description":"Move towards the bag, lower the gripper, and close the gripper to grasp the bag."},{"start_frame":26,"end_frame":80,"phase_type":"interact","description":"Lift the bag and move it toward the bottom drawer."},{"start_frame":81,"end_frame":100,"phase_type":"release","description":"Place the grasped bag in the drawer."}]}

[QUERY TEMPLATE]
Task: <language_instruction>
Detected phases: <json_serialized_phases>
\end{lstlisting}
\end{subfigure}

\vspace{0.4em}

\begin{subfigure}{0.98\textwidth}
\caption{Stage 1b: Bimanual task and object extraction from left/right phase streams}
\begin{lstlisting}[style=promptstyle]
[SYSTEM] You are a robot manipulation expert. Given a task instruction and detected gripper phases for a bimanual robot with separate left-arm and right-arm timelines, identify which object each arm interacts with, assign phases to the corresponding subtask, and generate phase descriptions. Each phase has start_frame, end_frame, and phase_type in {"grasp","interact","release","grasp_failure","null"}. Output a JSON list with keys: arm, object, start_location, target_location, action, tool_name_required, tool_usage_description, grasp_phases. Each entry must specify arm as left or right. Do not merge arms into one entry. If an arm has no valid interaction, emit an entry with an empty grasp_phases list.

[EXAMPLE]
Task: Pick up the red block with the left arm and place the blue cup with the right arm
Left-arm phases: [{"start_frame":0,"end_frame":20,"phase_type":"grasp"},{"start_frame":21,"end_frame":60,"phase_type":"interact"},{"start_frame":61,"end_frame":70,"phase_type":"release"}]
Right-arm phases: [{"start_frame":5,"end_frame":25,"phase_type":"grasp"},{"start_frame":26,"end_frame":65,"phase_type":"interact"},{"start_frame":66,"end_frame":75,"phase_type":"release"}]
Output: [{"arm":"left","object":"red block","start_location":null,"target_location":null,"action":"pick up","tool_name_required":null,"tool_usage_description":null,"grasp_phases":[{"start_frame":0,"end_frame":20,"phase_type":"grasp","description":"Move the left gripper to the red block and grasp it."},{"start_frame":21,"end_frame":60,"phase_type":"interact","description":"Lift the red block."},{"start_frame":61,"end_frame":70,"phase_type":"release","description":"Release the red block."}]},{"arm":"right","object":"blue cup","start_location":null,"target_location":null,"action":"place","tool_name_required":null,"tool_usage_description":null,"grasp_phases":[{"start_frame":5,"end_frame":25,"phase_type":"grasp","description":"Move the right gripper to the blue cup and grasp it."},{"start_frame":26,"end_frame":65,"phase_type":"interact","description":"Move the blue cup to the target location."},{"start_frame":66,"end_frame":75,"phase_type":"release","description":"Place the blue cup down."}]}]

[QUERY TEMPLATE]
Task: <language_instruction>
Left-arm phases: <json_serialized_left_phases>
Right-arm phases: <json_serialized_right_phases>
\end{lstlisting}
\end{subfigure}

\caption{Prompt templates used for Stage 1 task/object extraction in the annotation pipeline. The pipeline uses the single-arm prompt for standard trajectories and the bimanual prompt when separate left/right gripper phase streams are available.}
\label{fig:appendix_prompts_compact}
\end{figure*}

\end{document}